\documentclass[10pt]{article}
\usepackage[utf8]{inputenc}
\usepackage[margin=1in]{geometry}
\usepackage{amsmath,amssymb,amsfonts}
\usepackage{graphicx}
\usepackage{hyperref}
\usepackage{cite}
\usepackage{booktabs}
\usepackage{algorithm}
\usepackage{algorithmic}
\usepackage{microtype}
\usepackage{subcaption}
\usepackage{xcolor}
\usepackage{tabularx}
\usepackage{float}
\usepackage{caption}

\title{\vspace{0cm}\textbf{TerraBind}: \\ Fast and Accurate Binding Affinity Prediction \\ through Coarse Structural Representations}

\author{
    Matteo Rossi\textsuperscript{†} \hspace{1em}
    Ryan Pederson\textsuperscript{†} \hspace{1em}
    Miles Wang-Henderson \hspace{1em}
    Ben Kaufman \\[0.4em]
    Edward C. Williams \hspace{1em}
    Carl Underkoffler \hspace{1em}
    Owen Lewis Howell \hspace{1em}
    Adrian Layer \\[0.4em]
    Stephan Thaler \hspace{1em}
    Narbe Mardirossian \hspace{1em}
    John Anthony Parkhill
    \\[2em]
    \textit{Terray Therapeutics, Inc.} \\
    \textit{Monrovia, CA, USA}
}
\date{}

\begin{document}
\maketitle

\begingroup
    \renewcommand\thefootnote{\textdagger}
    \footnotetext{Equal contributions. Correspondence: \texttt{\{matteo.rossi,ryan.pederson\}@terraytx.com}}
\endgroup
\setcounter{footnote}{0}

\begin{abstract}
We present TerraBind, a foundation model for protein-ligand structure and binding affinity prediction that achieves 26-fold faster inference than state-of-the-art methods while improving affinity prediction accuracy by $\sim$20\%. Current deep learning approaches to structure-based drug design rely on expensive all-atom diffusion to generate 3D coordinates, creating inference bottlenecks that render large-scale compound screening computationally intractable. We challenge this paradigm with a critical hypothesis: full all-atom resolution is unnecessary for accurate small molecule pose and binding affinity prediction. TerraBind tests this hypothesis through a coarse pocket-level representation (protein C$_\beta$ atoms and ligand heavy atoms only) within a multimodal architecture combining COATI-3 molecular encodings and ESM-2 protein embeddings that learns rich structural representations, which are used in a diffusion-free optimization module for pose generation and a binding affinity likelihood prediction module. On structure prediction benchmarks (FoldBench, PoseBusters, Runs N' Poses), TerraBind matches diffusion-based baselines in ligand pose accuracy. Crucially, TerraBind outperforms Boltz-2 by $\sim$20\% in Pearson correlation for binding affinity prediction on both a public benchmark (CASP16) and a diverse proprietary dataset (18 biochemical/cell assays). We show that the affinity prediction module also provides well-calibrated affinity uncertainty estimates, addressing a critical gap in reliable compound prioritization for drug discovery. Furthermore, this module enables a continual learning framework and a hedged batch selection strategy that, in simulated drug discovery cycles, achieves 6$\times$ greater affinity improvement of selected molecules over greedy-based approaches. 
\end{abstract}


\section{Introduction}

Accelerating the discovery of small molecule therapeutics requires computational models that can balance high predictive accuracy with the throughput necessary to screen vast chemical spaces. Traditionally, structure-based drug design (SBDD) has relied on physics-based molecular docking methods~\cite{wang2016comprehensive} (e.g., AutoDock Vina~\cite{eberhardt2021autodock}, Glide~\cite{friesner2004glide}) which, while computationally efficient, often fail to capture induced-fit effects and suffer from limited scoring accuracy.

The recent emergence of deep learning models—such as AlphaFold2~\cite{jumper2021highly} and AlphaFold3~\cite{abramson2024accurate}, NeuralPLexer3~\cite{qiao2024neuralplexer3}, RoseTTAFold~\cite{baek2021accurate}, Boltz-1(x)~\cite{wohlwend2025boltz} and Boltz-2~\cite{passaro2025boltz}—has fundamentally shifted this landscape. These models jointly fold (co-fold) proteins and ligands, achieving high structural fidelity and binding affinity predictions that approach the accuracy of physics-based free energy calculations. However, this performance comes at a prohibitive computational cost. State-of-the-art systems like Boltz-2 rely on iterative diffusion processes to generate explicit 3D coordinates, requiring approximately 20 seconds per complex~\cite{passaro2025boltz}. This inference bottleneck renders them impractical for high-throughput binding affinity prediction in drug discovery contexts. For perspective, Terray's EMMI platform~\cite{terray2025emmi} generates over one billion unique experimental binding measurements per quarter, which underscores the need for computational methods that can operate at comparable or larger throughput.

Recent attempts to mitigate this latency, such as Boltzina~\cite{furui2025boltzina}, propose bypassing the structure module by reverting to docking-generated poses; however, this often reintroduces the rigid-pocket assumptions that co-folding was meant to solve. Furthermore, these architectures remain difficult to recalibrate when novel empirical evidence for a specific target becomes available. In contrast, cheaper structure-less \textit{sequence-based} models~\cite{wang2025pretrained,khokhlov2025drugform} clearly offer the throughput required for billions-scale virtual screening, but their performance is highly sensitive to the available training data. While we have found these models to be effective for well-characterized proteins within the EMMI platform, they routinely underperform structure-based counterparts on data-poor targets. This creates a fundamental tension in the field: traditional methods and sequence-based models are fast but often lack structural nuance or generalizability, while modern diffusion-based models are accurate but remain computationally intractable at the scale of industrial drug discovery.

Beyond computational bottlenecks, reliable deployment of binding affinity models is hindered by a lack of rigorous uncertainty quantification. While modern structure prediction architectures typically output well-calibrated confidence metrics (such as pLDDT and ipTM) to assess geometric plausibility, analogous mechanisms for binding affinity are notably absent. Critically, these structural confidence scores cannot simply be repurposed as proxies for affinity uncertainty; a model may generate a geometrically confident pose yet remain entirely uncertain about the magnitude of the binding interaction~\cite{passaro2025boltz}. Although uncertainty quantification models for potency have been proposed, they may lack target generality—often training only on select protein families like kinases—and have not been demonstrated within end-to-end co-folding workflows, relying instead on pre-existing experimental 3D structures~\cite{luo2023calibrated}.
Notably, the literature rarely addresses \textit{joint} uncertainty quantification for batches of molecules; however, we recently demonstrated that these methods facilitate more robust molecule selection and accelerate the iterative Design-Make-Test-Analyze (DMTA) cycles central to practical drug discovery~\cite{wang2025pretrained}.

\begin{figure}[H]
    \centering
    \includegraphics[width=\textwidth]{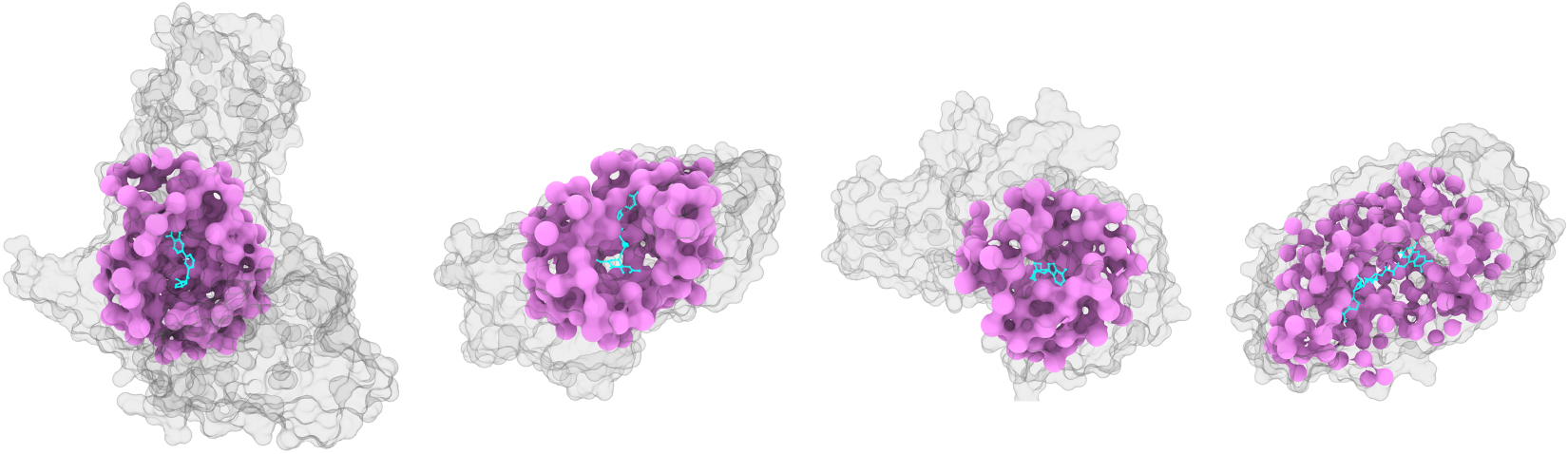}
    \caption{\textbf{TerraBind coarse structure prediction.} Example predicted protein-ligand complexes for PDB entries 8JJT, 8B1C, 8W9I, and 8RQH (left to right). The purple particles show the predicted binding site residues within 15\,\AA{} of the ligand, while the gray particles show the experimental ground truth structure. Proteins are represented by C$_\beta$ atoms only (no side chains), and ligands are shown with all heavy atoms. TerraBind predicts the local binding site structure rather than full protein co-folding.}
    \label{fig:coarse_structure}
\end{figure}

To simultaneously resolve these inference bottlenecks and reliability gaps, we first question the necessity of the computationally intensive components found in current co-folding architectures. We posit that the expensive diffusion-based generation used in state-of-the-art models is not only a bottleneck for throughput but is also largely superfluous for the core task of binding affinity prediction. This work is motivated by a critical hypothesis: \textit{full all-atom diffusion is not required to accurately predict binding affinity, nor is it strictly necessary for recovering high-fidelity ligand poses.} We propose that rich coarse-grained structural representations can capture the essential geometric and chemical information needed for both structure and binding affinity prediction without the overhead of explicit generative modeling (Fig.~\ref{fig:coarse_structure}).


To this end, we introduce \textbf{TerraBind} (Fig.~\ref{fig:performance_overview}), a foundation model for protein-ligand structure and binding affinity prediction designed for high-throughput applications—from virtual screening to hit optimization. Our approach focuses on learning rich structural representations at the binding site interface that directly inform binding affinity prediction, while employing lean model architectures—only $\sim$30M total trainable parameters compared to $\sim$509M for Boltz-2—to minimize computational cost without sacrificing accuracy. Our key contributions are:

\begin{itemize}

    \item A structure prediction module that combines pretrained encoders (COATI-3 for ligands~\cite{kaufman2024coati,kaufman2024latent}, ESM-2 for proteins~\cite{lin2023evolutionary}) with a lean 48-layer pairformer trunk architecture (27M parameters vs.\ 196M in Boltz-2 trunk) and a diffusion-free optimization scheme, achieving $\sim$26$\times$ faster inference than Boltz-2 while matching ligand pose accuracy on FoldBench, PoseBusters, and Runs N' Poses benchmarks.\footnote{Structure prediction is benchmarked against Boltz-1, which shares the same training data cutoff (September 30, 2021).\label{fn:cutoff}}

    \item A binding affinity prediction module that bypasses coordinate generation yet outperforms Boltz-2 on both proprietary and public benchmarks, achieving 20\% higher Pearson correlation across 18 drug discovery targets (surpassing on 15 of 18) and 16\% higher on CASP16.

    \item A built-in structural uncertainty score via pairwise distance entropy $H_{\text{LP}}$ (Eq.~\ref{eq:lp_entropy}), which correlates with both pose accuracy and binding affinity without requiring a separate confidence module.

    \item A 17\% affinity improvement on held-out compounds achieved by fine-tuning the structural Pairformer on a minimal set of proprietary crystals (6 and 3 structures for two targets).

    \item An epistemic neural network (epinet) module for calibrated affinity uncertainty, demonstrating that lower predicted uncertainty correlates with higher prediction success rates.

    \item A continual learning framework that leverages joint samples from the epinet module for batch selection, achieving 6$\times$ greater affinity improvement than greedy strategies in simulated DMTA cycles.
    
\end{itemize}

\begin{figure}[H]
    \centering
    \begin{subfigure}[b]{0.48\textwidth}
        \centering
        \includegraphics[width=\linewidth]{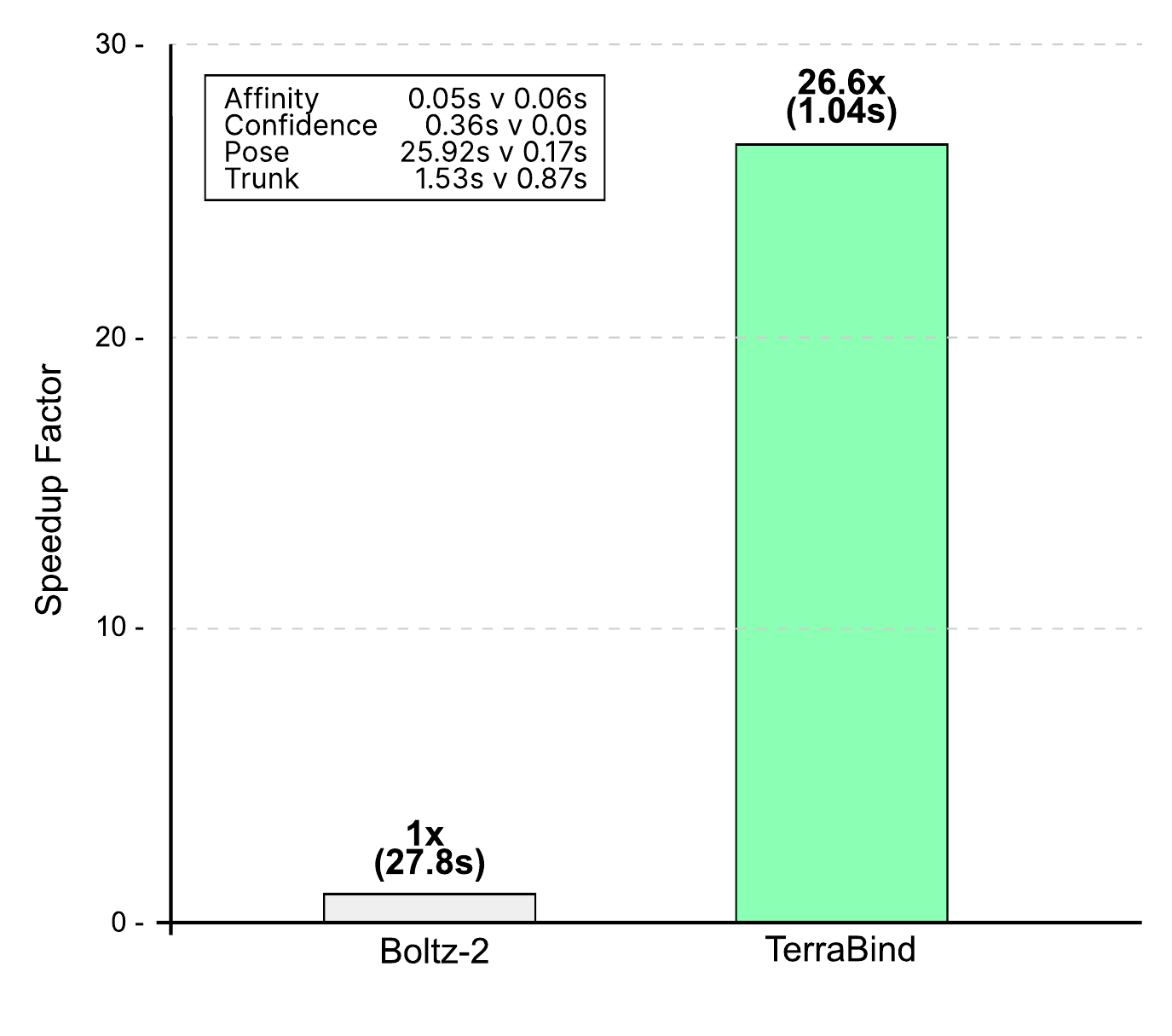}
        \caption{Inference throughput ($26\times$ speedup).}
        \label{fig:speedup}
    \end{subfigure}
    \hfill
    \begin{subfigure}[b]{0.48\textwidth}
        \centering
        \includegraphics[width=\linewidth]{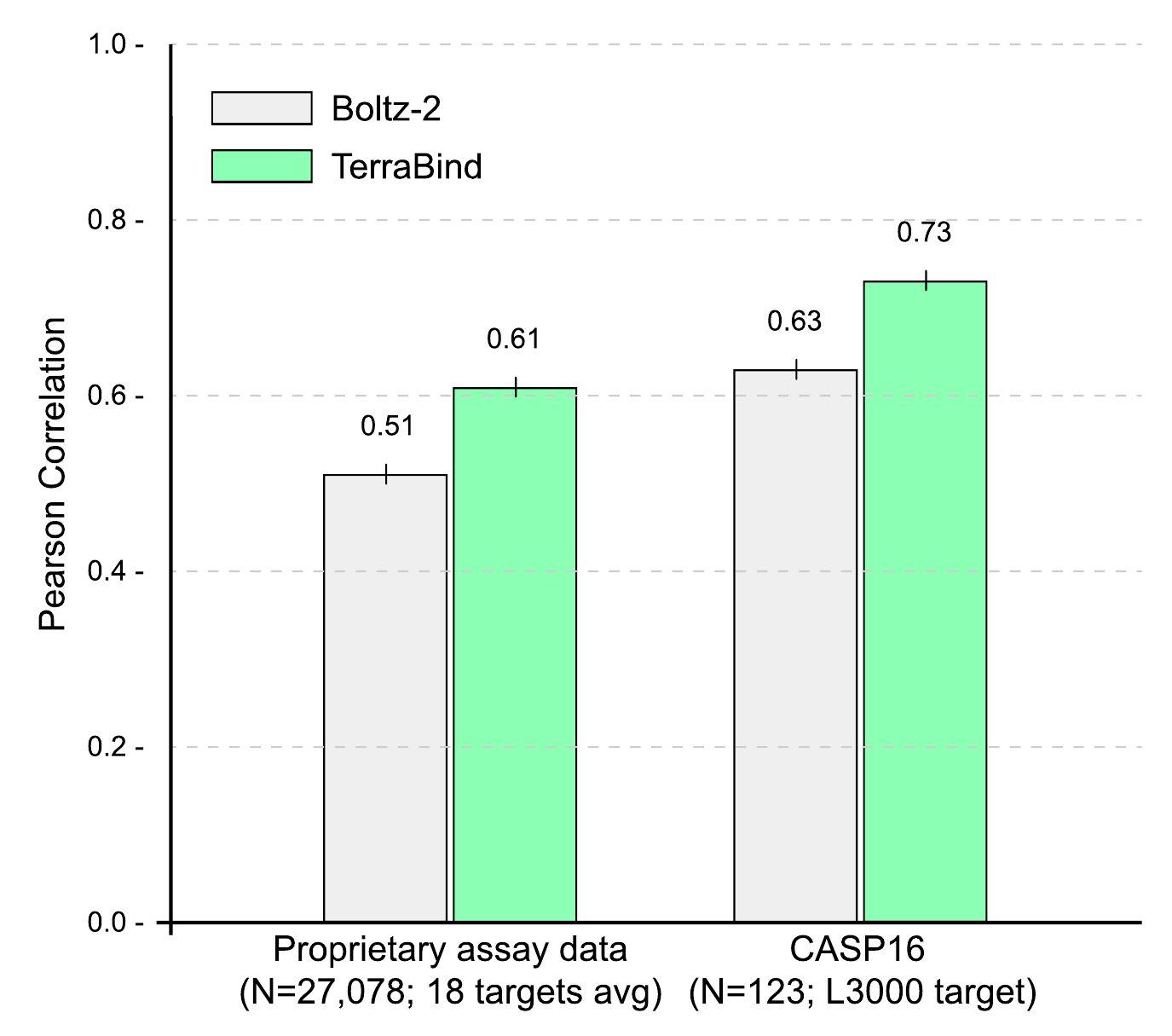} 
        \caption{Binding affinity correlation.}
        \label{fig:affinity_intro}
    \end{subfigure}
    \caption{\textbf{TerraBind performance overview.} 
        (a)~End-to-end inference time per complex on a single A6000 GPU (196 tokens, 10 samples), demonstrating a $26\times$ speedup over Boltz-2.
        (b)~Binding affinity prediction (Pearson correlation) on CASP16 and proprietary assay data, showing up to 20\% improvement.}
    \label{fig:performance_overview}
\end{figure}

\begin{figure}[H]\ContinuedFloat
    \centering
    \begin{subfigure}[b]{0.78\textwidth}
        \centering
        \includegraphics[width=\linewidth]{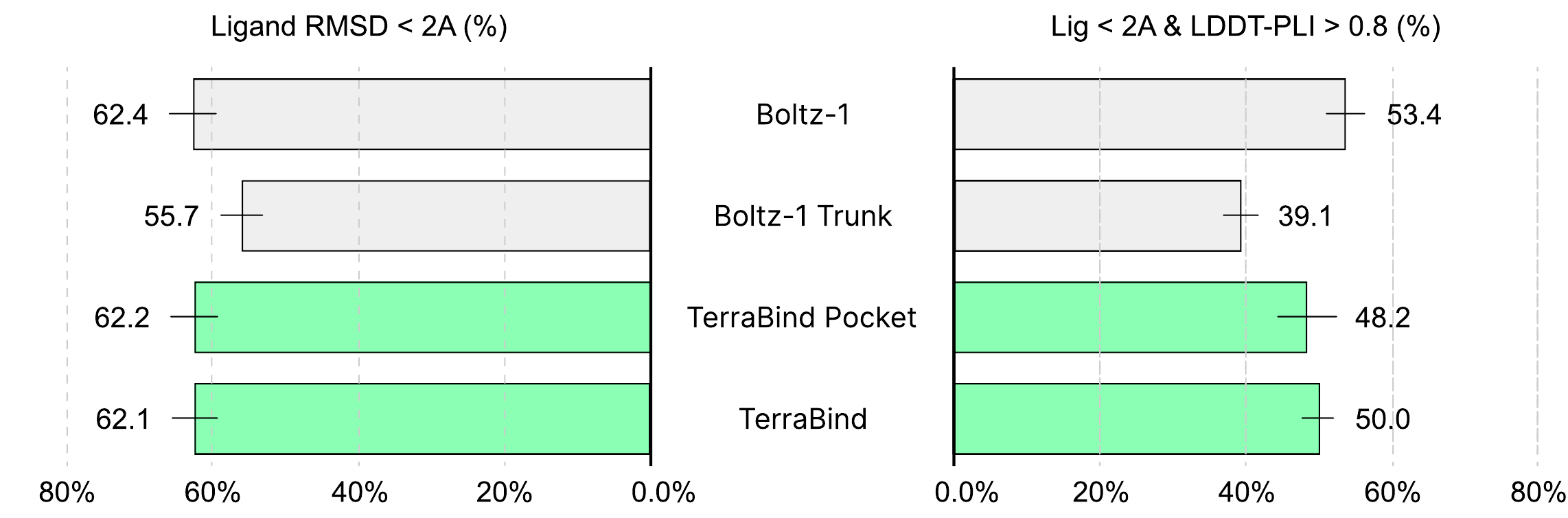}
        \caption{Structure prediction success rates.}
        \label{fig:structure}
    \end{subfigure}
    \caption{\textbf{TerraBind performance overview (cont.).} 
        (c)~Structure prediction performance compared to Boltz-1\textsuperscript{\ref{fn:cutoff}}, aggregated across FoldBench, PoseBusters, and Runs N' Poses benchmarks. Left bars show ligand RMSD~$<2$\AA\ success rate; right bars show a stricter metric (RMSD~$<2$\AA\ and LDDT-PLI~$>0.8$) that captures binding-relevant geometry. Models: Boltz-1 (full diffusion pipeline), Boltz-1 Trunk (Boltz-1 trunk representation with our coordinate optimization), TerraBind Pocket (196-token pocket context), and TerraBind (full protein context). See Section~\ref{sec:structure_benchmarks} for detailed model descriptions.}
\end{figure}

\section{Methods}

\begin{figure*}[ht!]
    \centering
    \includegraphics[width=1.0\linewidth]{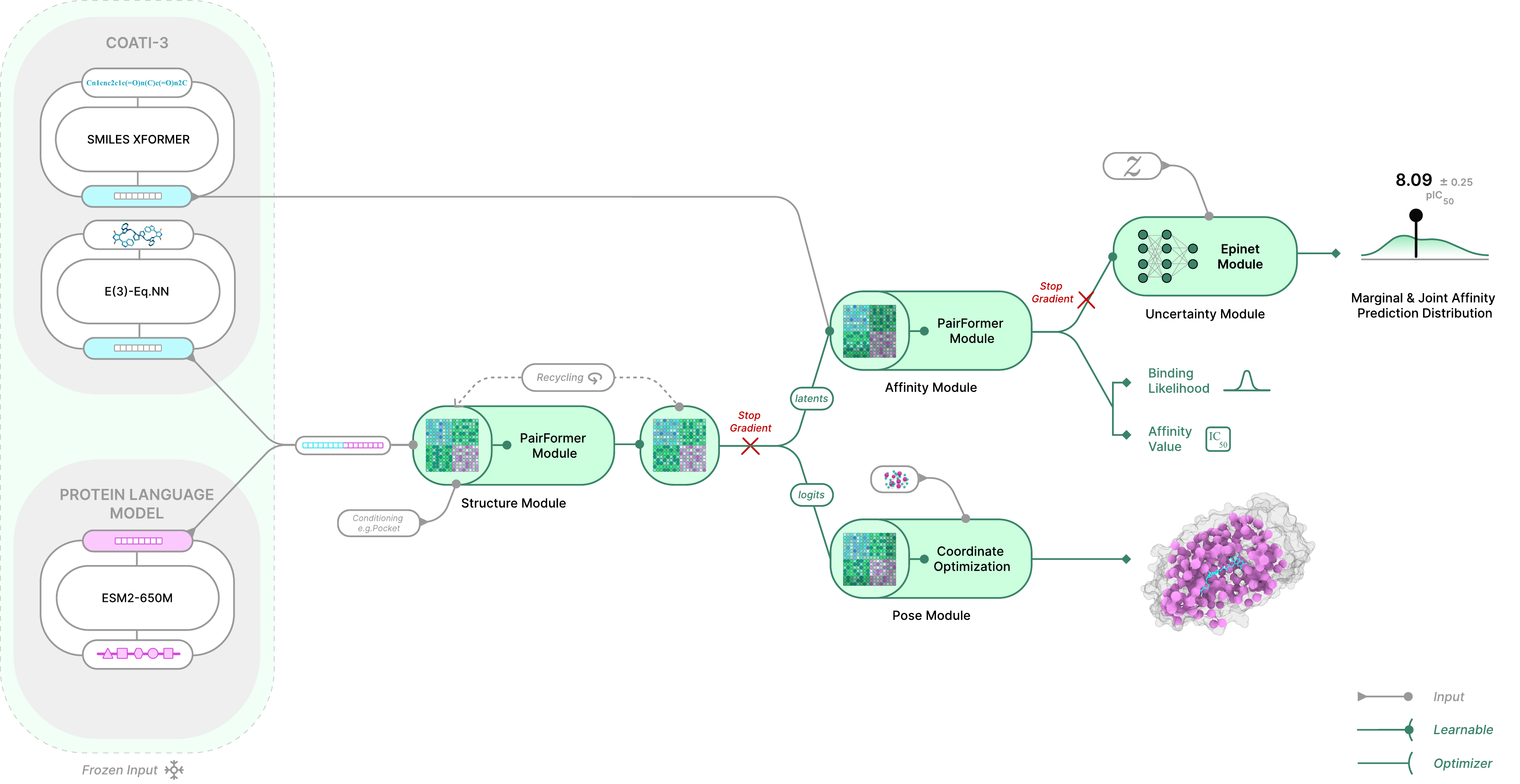}
    \caption{\textbf{TerraBind Architecture.} The model consists of four main components: (1) \textit{Frozen pretrained encoders}: COATI-3 (E(3)-equivariant encoder and SMILES transformer for ligands) and ESM-2 (language model for proteins) provide initial representations without requiring MSA generation at inference time. (2) \textit{Structure module}: A 48-layer pairformer architecture with triangle attention and multiplication learns pocket-level structural representations by predicting categorical distributions over distance bins for all token pairs. The pairwise distance entropy provides built-in structural uncertainty quantification. (3) \textit{Pose module}: For coordinate generation when needed, simple coarse-grained optimization from pairformer distance logits produces 3D structures in $<$0.2 second per complex without requiring diffusion. (4) \textit{Affinity module}: A specialized 6-layer pairformer operates on frozen structural features to predict binding likelihood (binary classification) and affinity values (continuous regression), with an epistemic neural network (epinet) providing calibrated uncertainty quantification.}
    \label{fig:architecture}
\end{figure*}

The TerraBind architecture consists of four main components illustrated in Figure~\ref{fig:architecture}: (1) frozen pretrained encoders (COATI-3 and ESM-2) that provide initial chemical and biological representations, (2) a structure module with a pairformer trunk (hereafter PF) that learns pocket-level structural representations through distance prediction, with pairwise distance entropy providing built-in structural uncertainty, (3) a pose module that generates 3D coordinates via coarse-grained optimization, and (4) an affinity module that maps structural features to binding strength, with an epistemic neural network providing calibrated affinity uncertainty. We describe each component in detail below.

\subsection{Frozen Pretrained Encoders}

TerraBind employs frozen pretrained encoders that provide strong initial representations without requiring expensive multiple sequence alignment (MSA) generation at inference time.

\textbf{COATI-3 Small-molecule Encoder:} COATI-3 is a multimodal small-molecule encoder. Like previous COATI model iterations~\cite{kaufman2024coati,kaufman2024latent}, COATI-3 is trained using a contrastive loss but utilizes $3$ chemical modalities: SMILES string, 2D graph, and 3D conformer point cloud (COATI-1 and COATI-2 did not include a 2D graph representation). Training datasets include those in COATI-1~\cite{kaufman2024coati} and COATI-2~\cite{kaufman2024latent}, but also an expanded sampling of molecules from 
PCCL~\cite{Bedart2024}, SAVI~\cite{Patel2020}, BindingDB~\cite{gilson2016bindingdb}, Plinder~\cite{Durairaj2024.07.17.603955}, COCONUT~\cite{Sorokina2021}, Buchwald-Hartwig~\cite{Ahneman2018} data, and the USPTO~\cite{Lowe2012}, 
culminating in a training dataset size of over a billion compounds.

The SMILES encoder is a standard text transformer architecture~\cite{kaufman2024latent}, the 2D graph encoder is a graph transformer model, and the 3D point encoder is an equivariant neural network based on the Allegro~\cite{musaelian2023learning,geiger2022e3nn} model architecture. See Ref.~\cite{kaufman2024latent} for further details on the contrastive training approach. The contrastive loss considers aggregated $768$-dimensional vector embeddings across the $3$ modalities. The 2D graph and 3D point encoders create atom-level vector embeddings of the molecule before a final aggregation step to produce the $768$-dimensional molecule embedding, thus making such intermediate atom-level embeddings an attractive starting point for structure prediction models. Additionally, the chemical space exposed to the COATI-3 pretraining is vast relative to the chemical space accessible to structural training, which spans fewer than 50,000 unique ligands.

\textbf{Protein Encoder - ESM-2 (650M):} A frozen pretrained masked language model trained on $\sim$65M unique protein sequences~\cite{lin2023evolutionary}. ESM-2 captures evolutionary information and implicit structural propensities directly from sequence without requiring MSA generation. As with the ligand encoder, the sequence space exposed to ESM-2 pretraining is vast relative to structural training data; starting from pretrained residue embeddings provides more efficient training and more robust global sequence context than embeddings learned on-the-fly.


Both ligand and protein encoders remain frozen throughout all training stages to preserve their generalization capabilities to chemical and biological spaces beyond the PDB while reducing computational cost and memory requirements.

\subsection{Structure Prediction Module}

The structure prediction module learns to predict 3D geometric relationships between atoms in protein-ligand complexes. Starting from frozen pretrained encoder outputs, the module constructs pair representations and processes them through a trainable pairformer trunk to predict categorical distributions over pairwise distance bins, trained with cross-entropy loss (see Fig.~\ref{fig:architecture}).

\subsubsection{Pairformer Trunk}

The pairformer trunk architecture in TerraBind is adapted from Ref.~\cite{abramson2024accurate}. The core operations—triangle attention (Appendix Eq.~\ref{eq:triangle_attention}) and triangle multiplication (Appendix Eq.~\ref{eq:triangle_mult})—update pair representations by propagating information along edges of an implicit graph. The factorizable input pair representations are derived from ESM-2 protein residue representations and COATI-3 ligand atom representations from the $E(3)$-equivariant Allegro encoder, rather than trainable atom encoder-based modules. Multiple sequence alignment (MSA) input features are not utilized in our model, and the single sequence representation component of the PF has been removed; these representations are utilized in the downstream diffusion module in Ref.~\cite{abramson2024accurate}, which is absent in our model. By eliminating single representations, we reduce the structure module parameter count by $\sim$5$\times$ from 147M to 27M for a 48-layer pairformer.

Following Ref.~\cite{wohlwend2025boltz}, we project the final pair representations into 64 pairwise distance bins: 62 bins evenly spaced from $2\text{\AA}$ to $22\text{\AA}$, plus boundary bins for covalent-range distances ($<2\text{\AA}$) and long-range interactions ($>22\text{\AA}$). Distances are predicted between all ligand heavy atoms and protein residue centers (represented by C$_\beta$ atoms, or C$_\alpha$ for glycine). The PF produces a distogram of pairwise distance bin probability distributions for all system pairs. The model is trained using a categorical cross-entropy loss over these distance bins, weighted by pair type. Pair types are distinguished by interaction class (intra- vs. inter-molecular) and entity type (ligand-ligand, protein-protein, ligand-protein), each weighted differently to emphasize binding-relevant geometry (see Appendix~\ref{sec:structure_loss} for the full loss formulation).
From these bin distributions, we derive two useful quantities: the expected pairwise distance and the normalized pairwise entropy (see Appendix~\ref{sec:expected_distance} and \ref{sec:pairwise_entropy} for equations). The expected distance is computed as the probability-weighted average of bin centers. These expected distances are used to define a binding pocket around the ligand—specifically, protein residues within $15\text{\AA}$ of any ligand atom. This pocket definition serves as input context for both the affinity module (Section~\ref{sec:affinity_module}) and the pose module (Section~\ref{sec:pose_module}), which performs coordinate optimization within this $15\text{\AA}$ cutoff.
The normalized pairwise entropy is computed from the predicted probability distribution and scaled to lie between 0 and 1. For assessing binding confidence, we average these pairwise entropies over ligand-protein pairs to obtain a single interface entropy score, $H_{\text{LP}}$.\footnote{Throughout this paper, subscripts LP, LL, and PP denote ligand-protein, ligand-ligand, and protein-protein pairs, respectively.} This score serves two purposes: (1) as an uncertainty estimate for the predicted binding geometry, where high entropy indicates high uncertainty, and (2) as a zero-shot binding affinity signal, where lower entropy correlates with stronger binding.

\subsubsection{Structure Module Training Data}

Our structure prediction training leverages two primary data sources to maximize diversity while maintaining quality. Across all datasets, we filter to retain only protein and small molecule complexes, excluding systems containing nucleic acids (DNA/RNA), single ion ligands, or other non-drug-like entities.

\textbf{Experimental structures:} All structures from the Protein Data Bank (PDB)~\cite{wwpdb2019protein} released before 2021-09-30 (167,588 structures after filtering). This encompasses diverse structural biology data to establish broad structure learning capabilities.

\textbf{Distillation data:} To increase training data size and diversity, we employ distillation from state-of-the-art structure prediction models trained before the PDB cutoff date to prevent data leakage:

\begin{itemize}
\item \textbf{AlphaFold Database (AFDB):} High-confidence AlphaFold2 predictions on single-chain monomers from Swiss-Prot~\cite{jumper2021highly,varadi2024alphafold}, filtered to $\text{pTM} > 0.9$ to ensure structural quality (542,378 structures). This exposes the model to protein sequence space beyond experimentally determined structures.

\item \textbf{BindingDB:} Computationally generated protein-ligand complexes from Boltz-1x~\cite{wohlwend2025boltz} predictions for complexes in BindingDB~\cite{liu2007bindingdb}, filtered to $\text{pIC50} > 6$ and $\text{ipTM} > 0.9$ (438,957 complexes). These predictions provide additional protein-ligand binding mode diversity beyond experimental structures.
\end{itemize}

Since TerraBind does not include a trainable all-atom structural refinement module downstream of the pairformer trunk (e.g., a diffusion module~\cite{abramson2024accurate}), distillation data is especially critical in our approach to ensure high-quality distance predictions at the pairformer level.

\subsubsection{Structure Module Training Protocol}

We employ a three-stage curriculum (see Appendix Table~\ref{tab:training_stages}) designed to transition from global structural learning to focused binding site geometry. All stages use an effective batch size of 128 distributed across 4 NVIDIA H100 nodes (32 GPUs total).

\begin{itemize}
\item \textbf{Stage 1} (70k steps): Pretrains on diverse interface types—including protein-protein, protein-ligand, ligand-ligand, and apo structures—using 384-token crops from PDB, AFDB, and BindingDB. This establishes broad structural priors across interaction types.

\item \textbf{Stage 2} (20k steps): Shifts to ligand-centered 256-token crops focused on protein-ligand interfaces, using only PDB and BindingDB data. Intra-ligand ($2\times$) and ligand-protein ($5\times$) pairwise distance losses are upweighted to emphasize binding interaction geometry.

\item \textbf{Stage 3} (15k steps): Continues with ligand-centered 256-token crops on protein-ligand interfaces, but fine-tunes exclusively on experimental PDB structures with equal loss weights to prioritize ground truth fidelity.
\end{itemize}

Total training completes in 105k steps. To manage computational cost, we employ a conservative bfloat16 (bf16) mixed-precision strategy. Computationally intensive trunk operations (e.g., triangle attention and multiplication, see Appendix~\ref{sec:pairformer_arch}) via NVIDIA cuEquivariance kernels~\cite{cuequivariance} use bf16, while numerically sensitive components (losses, distance projections) and all model weights remain in full fp32 precision to ensure stability. Combined with our compact 256-token crops (compared to 512+ tokens in comparable methods~\cite{abramson2024accurate,passaro2025boltz}), the absence of a diffusion module, and no separate confidence module to train, these choices yield approximately $2\times$ reduction in training time relative to comparable methods.

The transition to smaller crops is motivated by the realities of small-molecule drug discovery. 
Typically, for a given drug campaign, there are specific binding site(s) of interest. 
Training the model with limited context better enables inference with limited context surrounding the binding site of interest, which yields cubical speedups versus full protein sequence inference without significantly sacrificing accuracy.   
In practice, we observe that relevant small molecule binding pocket contexts (number of heavy ligand atoms plus protein residues within 15$\text{\AA}$ of any ligand atom) rarely exceed 200 tokens (see Appendix Figure~\ref{fig:affinity_15A_context}), validating our 256-token crop size during training.

A key distinction of our approach is the emphasis on learning rich representations at the pairformer level. Existing co-folding methods typically rely on diffusion modules for all-atom structural refinement, as training a pairformer to predict full protein structures with all-atom fidelity is computationally intractable. In contrast, by focusing exclusively on small-molecule binding and operating with compact crops centered on the binding site, we can train the pairformer to capture all binding-relevant geometric information directly. This design choice is critical because in TerraBind, the pairformer representations directly inform both pose generation and affinity prediction—there is no downstream diffusion module to compensate for representational deficiencies.

\subsubsection{Structure Prediction Benchmarks}
\label{sec:structure_benchmarks}

We compare TerraBind against Boltz-1~\cite{wohlwend2025boltz} for structure prediction, as both models share the same training data cutoff (September 30, 2021), enabling fair comparison. We present four model configurations that isolate different components of the pipeline:

\begin{itemize}
\item \textbf{TerraBind} and \textbf{Boltz-1 Trunk} both apply identical coordinate optimization to generate poses, differing only in the underlying trunk representations—TerraBind's pairformer versus Boltz-1's trunk module. This controlled comparison directly evaluates the quality of learned representations, which is the critical factor for downstream affinity prediction.

\item \textbf{TerraBind Pocket} extends TerraBind with a two-stage strategy: initial inference on the full protein to identify the binding pocket, followed by cropping to a compact 196-token context and refined prediction on the local environment (see Appendix~\ref{sec:pocket_crop_algorithm} for the pocket cropping algorithm). Since pairformer computation scales $O(N^3)$ with token count, this pocket-focused inference substantially reduces computational cost for large proteins, enabling high-throughput binding affinity prediction across millions of compounds. Notably, TerraBind Pocket maintains robust performance despite the reduced context.

\item \textbf{Boltz-1} uses Boltz-1's full diffusion-based structure generation module, representing the complete Boltz-1 pipeline.
\end{itemize}

For all methods, we generate 10 pose samples and select the best pose for evaluation. For optimization-based methods (TerraBind, TerraBind Pocket, and Boltz-1 Trunk), we select the pose with the lowest optimization loss. For diffusion-based methods (Boltz-1), we select the pose with the highest ipTM confidence score.

We evaluate structure prediction performance across four complementary benchmarks that test different aspects of generalization. For all benchmarks, we report two metrics: (1) ligand RMSD $<2\text{\AA}$ success rate, and (2) a combined success rate requiring both ligand RMSD $<2\text{\AA}$ and LDDT-PLI $>0.8$. Across all datasets, we curate and filter to retain only protein-small molecule complexes, excluding systems containing nucleic acids (DNA/RNA), single ion ligands, or other non-drug-like entities.

\begin{itemize}
\item \textbf{FoldBench} ($n=556$): A comprehensive low-homology benchmark derived from PDB structures deposited after the AlphaFold~3 validation cutoff (January 2023), with targets filtered to remove high sequence and structural similarity to training data~\cite{xu2025foldbench}.

\item \textbf{PoseBusters} ($n=307$): A curated set of high-quality, drug-like protein-ligand complexes released after 2021~\cite{buttenschoen2024posebusters}. The benchmark was designed specifically to assess generalization of docking and co-folding methods to novel structures.

\item \textbf{Runs N' Poses} ($n=2{,}687$): A zero-shot co-folding benchmark comprising high-resolution protein-ligand systems released after the common training cutoff of September 30, 2021~\cite{vskrinjar2025have}.

\item \textbf{Proprietary crystal structures}: A Terray proprietary validation set of experimentally determined protein-ligand co-crystal structures from active drug discovery programs. These structures provide an unbiased assessment of prospective prediction accuracy on real-world therapeutic targets not available in public databases.
\end{itemize}

\subsection{Coarse-grained Pose Module}
\label{sec:pose_module}

The distogram representation produced by the pairformer is inherently coarse-grained: only protein residue centers (C$_\beta$ atoms) and ligand heavy atoms are represented, with all other residue atoms neglected. To evaluate ligand pose quality using familiar metrics such as RMSD, we employ a simple optimization routine to generate 3D point clouds consistent with the predicted pairwise distances (see Figure~\ref{fig:coarse_structure} for an example). Importantly, this module involves no learned parameters—it is purely an optimization procedure used for pose visualization and structural representation evaluation.

Specifically, we initialize a 3D point cloud from random noise and optimize coordinates to minimize the discrepancy between the resulting pairwise distances and the expected distances derived from the pairformer's predicted bin distributions (see Appendix~\ref{sec:pose_module_details} for the full optimization algorithm). The optimization is restricted to ligand heavy atoms and protein residue centers within $15\text{\AA}$ of the ligand, matching the pocket definition used elsewhere in the pipeline. This routine typically converges in a few hundred steps and can be parallelized across large batch sizes to generate multiple pose samples efficiently, making it far more efficient than parameter-heavy diffusion models. We generate multiple samples with different random initializations; at convergence, each sample yields a different optimization loss, and we select the pose with the lowest loss as the best candidate.

This approach resembles an implicit energy-based model paradigm, which has shown promise over traditional diffusion and flow-based generative models~\cite{wang2025equilibrium,roney2025protein}. A key distinction from other co-folding methods is that these generated 3D coordinates are \textit{not} passed to the affinity module. In TerraBind, affinity prediction operates directly on the pairformer's latent representations and distance distributions, bypassing explicit coordinate generation entirely. The pose module serves solely to produce interpretable structures for visualization and to evaluate the quality of the learned structural representations.

\subsection{Binding Affinity Prediction Module}

The binding affinity prediction module learns to map pretrained structural representations to binding affinity. End-to-end binding affinity prediction training features a stop-gradient in front of the pairformer, enabling modular training and deployment of affinity modules. In practice, we match the affinity module to the pairformer it was trained with. We also introduce an affinity likelihood module for uncertainty quantification and molecule selection.


\subsubsection{Affinity Module Architecture}
\label{sec:affinity_module}

The affinity module is a six-layer pairformer architecturally identical to the structural pairformer, operating on frozen (stop-gradient) inputs. The inputs consist of: (1) pairwise structural pairformer latents (128-dimensional), (2) pairwise distance bin probabilities (64-dimensional), (3) per-ligand-atom COATI-3 embeddings (768-dimensional), (4) per-residue ESM-2 embeddings (1280-dimensional), and (5) an aggregated ligand COATI-3 embedding (768-dimensional).

A series of pair conditioning layers project these inputs to a 128-dimensional pair representation. Unlike Boltz-2, which conditions on distograms derived from diffused 3D coordinates~\cite{passaro2025boltz}, we condition directly on the pairwise distance bin probabilities from the structural pairformer. The aggregated ligand COATI-3 embedding is broadcast to ligand-ligand pairs only, capturing global ligand context. After the six-layer affinity pairformer, pair latents are mean-pooled to produce a single complex representation $\mathbf{g}$, which is passed through MLP layers to two prediction heads:
\begin{equation}
p_{\text{bind}} = \sigma(f_{\text{cls}}(\mathbf{g})) \in [0,1], \qquad \hat{y} = f_{\text{reg}}(\mathbf{g}) \in \mathbb{R}
\label{eq:affinity_heads}
\end{equation}
where $p_{\text{bind}}$ is the binding likelihood (binary classification) and $\hat{y}$ is the predicted binding affinity in $\log_{10}$ units.

We precompute frozen structural pairformer representations for all training examples, retaining only protein residues within $15\text{\AA}$ of any ligand atom based on expected pairwise distances (Eq.~\ref{eq:pocket_definition}). Protein-protein pairs are masked during the affinity pairformer forward pass. This reduces storage and accelerates training by focusing on binding-relevant representations. Across our dataset, the number of ligand heavy atoms plus pocket residues rarely exceeds 200 tokens (see Figure~\ref{fig:affinity_15A_context}).

\subsubsection{Affinity Likelihood Module}
\label{sec:affinity_likelihood_module}

We utilize an epistemic neural network (epinet) architecture~\cite{osband2023epistemic,wang2025pretrained} to provide quantitative estimates of posterior affinity likelihood. In Ref.~\cite{wang2025pretrained} we introduced the application of the epinet architecture to binding affinity prediction. The epinet is a simple and small MLP architecture which takes the frozen pretrained affinity module complex latent $\mathbf{g}$ (see previous section) and an epistemic index embedding $\mathbf{z} \sim \mathcal{N}(\mathbf{0}, \mathbf{I}_{256})$ sampled from unit Gaussian to produce a predicted residual $r_\theta(\mathbf{g}, \mathbf{z})$:
\begin{equation}
\hat{y} \sim \hat{y}_{\text{TB}} + r_\theta(\mathbf{g}, \mathbf{z})
\label{eq:epinet_prediction}
\end{equation}
where $\hat{y}_{\text{TB}}$ is the base affinity module quantitative affinity prediction, and $r_\theta(\mathbf{g}, \mathbf{z}) = f_\theta(\mathbf{g}, \mathbf{z}) + f_\phi(\mathbf{g}, \mathbf{z}) $ is the learned residual consisting of a trainable component $f_\theta$ and a frozen prior network $f_\phi$~\cite{osband2023epistemic,wang2025pretrained}. In this work, the prior network is simply a frozen, randomly initialized small MLP network; however, specialized pretrained prior networks may also be used. In our recent work~\cite{wang2025pretrained}, we find that pretraining a prior network using a reference process to generate synthetic datasets with non-Gaussian marginals can better capture the bounded, skewed distributions observed in real assays within practical drug discovery.

During training, the epinet model learns the affinity module residual for training data irrespective of the epinet index. At inference time, sampling multiple indices $\{\mathbf{z}_i\}_{i=1}^{K}$ and applying Eq.~\ref{eq:epinet_prediction} yields samples from the binding affinity posterior $p(y \mid \mathbf{g})$, as well as the joint posterior between $N$ input complexes,  $p(y_{1:N} \mid \mathbf{g}_{1:N})$. In practice, we often observe converged statistics with approximately $1000$ epinet index samples. Unlike large ensemble models, there is negligible overhead in sampling repeatedly from this epinet architecture. The expensive bottleneck is still overwhelmingly the structural pairformer and affinity pairformer modules that produce the latent representation $\mathbf{g}$, which does not need to be recomputed for each epinet sample.

These epinet sample statistics (such as sample standard deviation or interquartile range) can be used to quantify the uncertainty in the prediction. For instance, protein-ligand complexes nearby to training data will tend to have binding affinity distributions with low standard deviation, reflecting higher certainty in binding affinity prediction.
Importantly, the \textit{joint} posterior between pairs of inputs can also be leveraged in batched optimization schemes. Our recent findings~\cite{wang2025pretrained} show that utilizing epinet-based joint predictive distributions allows us to model the correlation structure required to mediate the trade-off between maximizing molecular objectives and hedging selection risks within a limited batch size, $B$.
For the task of maximizing binding affinity, we found the following EMAX (expected maximum value in a batch) acquisition function to outperform traditional greedy-based approaches used in drug discovery:
\begin{equation}
    \mathbb{E}_{\hat{y}_{1:B} \sim p(y_{1:B})}[\text{max}(\hat{y}_{1:B})] \, .
\label{eq:emax}
\end{equation}
Furthermore, in Section~\ref{sec:continual_learning_details}, we implement a \textit{continual learning} scheme to quickly incorporate real-world observations into the model predictions without tedious and computationally expensive retraining.

\subsubsection{Binding Affinity Training Data}

Our affinity training data focuses on high-quality binding affinity assays with careful curation to minimize noise while supporting both hit discovery (binary classification) and hit-to-lead optimization (quantitative affinity). We aggregate data from ChEMBL~\cite{zdrazil2024chembl}, BindingDB~\cite{liu2007bindingdb}, PubChem~\cite{kim2023pubchem} high-throughput screening (HTS) binary data only for curated assays in the MF-PCBA dataset~\cite{buterez2023mf}, and binary data from a fragment screening dataset from CeMM~\cite{offensperger2024large}. We also generate ${\sim}1.2$M synthetic decoy compounds which are treated as binary data. Here, we sample ChEMBL and BindingDB binder compounds, assume that they are target-selective, and assign them as non-binders for unrelated target sequences in the training data pool~\cite{passaro2025boltz}. Note that this resulting training dataset is only a rough subset of the one used in Boltz-2, which includes additional assay sources and PubChem assays.

We employ a structure-uncertainty prefiltering strategy, using the ligand-protein entropy $H_{\text{LP}}$ (Eq.~\ref{eq:lp_entropy}) to identify structurally unreliable data. Dataset entries with potency $< 1\mu\text{M}$ and $H_{\text{LP}} > 0.7$ are excluded. We do not differentiate between different affinity measurements ($K_i$, $K_d$, IC50, EC50, etc.); all values are transformed to $\log_{10}$ scale.

\subsubsection{Affinity Module Training Procedure}
\label{sec:affinity_module_training_procedure}

We train the affinity module using precomputed structure embeddings and do not update structure model weights. We supervise binary classification and quantitative binding affinity jointly, where the total loss is a sum of these two loss contributions~\cite{passaro2025boltz}. For binary classification, we employ a focal loss~\cite{lin2017focal} to deal with vast class imbalances in the binary data (often many more negatives than positives in HTS data). For quantitative binding affinity, a Huber loss~\cite{huber1964robust} is used for continuous regression in the \textit{absolute} case for standardized target affinity values, as well as (with higher loss weight) a \textit{relative} case, where the target values are the pairwise intra-assay differences; see Ref.~\cite{passaro2025boltz} for details. The Huber loss mitigates the effect of experimental and data gathering noise, as it is less sensitive to small prediction errors and data outliers. The relative loss component is crucial when training on data stemming from a wide array of different assays: due to different assay conditions and types, data from different assay sources is typically not directly comparable, especially given very limited detailed assay information~\cite{landrum2024combining}.

To achieve the desired effects of these losses, we employ a custom affinity batch sampler. For quantitative affinity assays, we employ a simplified approach versus Boltz-2, giving uniform sampling probability across all assays and complexes (instead of pre-assigning sampling rates based on the dynamic range of reported compound affinities in the assay). During training, an assay is sampled, and 5 random complexes from that assay are sampled into the training batch. For binary classification assays, an assay is sampled, and 1 random positive complex and 4 random negative complexes are sampled from that assay.

\subsubsection{Affinity Likelihood Module Training Procedure}

The affinity likelihood module utilizes a pretrained affinity module that is kept frozen. Here, we supervise on quantitative binding affinity data only and do not consider binary training data. We use the same affinity batch sampling approach as in affinity module training; however, we forgo the relative loss component. For each training step, a single epistemic index embedding $\mathbf{z} \sim \mathcal{N}(\mathbf{0}, \mathbf{I}_{256})$ is sampled from standard Gaussian noise and utilized across all examples in the current training batch. The training loss is defined as:
\begin{equation}
\mathcal{L}_{\text{epinet}} = \text{Huber}\left( y,\; \hat{y}_{\text{TB}} + r_\theta(\text{sg}(\mathbf{g}), \mathbf{z});\; \delta = 0.5 \right)
\label{eq:epinet_loss}
\end{equation}
where $y$ is the ground truth affinity, $\hat{y}_{\text{TB}}$ is the affinity prediction from the frozen TerraBind affinity model, $r_\theta$ is the predicted residual parameterized by the epinet weights $\theta$, and the operator $\text{sg}(\cdot)$ indicates a stop-gradient, ensuring that the epinet objective does not back-propagate into the base TerraBind affinity latent representation $\mathbf{g}$.

\subsubsection{Affinity Prediction Benchmarks}

We evaluate binding affinity prediction on two benchmark categories that test different aspects of model generalization: a public challenge dataset and diverse proprietary assays from active drug discovery programs.

\begin{itemize}
\item \textbf{CASP16:} Binding affinity prediction challenge with experimental validation on two recent protein-ligand complexes (L1000 and L3000 targets)~\cite{zhang2025assessment}. These challenges emphasize quantitative affinity prediction accuracy.

\item \textbf{Proprietary assay data:} Terray validation set from active drug discovery programs covering 18 diverse targets and associated chemical series pursued in realistic small molecule drug campaigns. These assays test model performance on prospective predictions and domain-specific challenges. Each target assay evaluated here contains more than 95 IC50 data points.
\end{itemize}


\section{Results}

\subsection{Structure Prediction Performance}
\label{sec:structure_performance}

We evaluate TerraBind's structure prediction capabilities across three public benchmarks—FoldBench, PoseBusters, and Runs N' Poses—as well as proprietary crystal structures. For all methods, we generate 10 pose samples per complex and apply symmetry correction to all final poses. For optimization-based methods (TerraBind, TerraBind Pocket, Boltz-1 Trunk), we select the best pose by lowest optimization loss; for diffusion-based methods (Boltz-1), we select by highest ipTM confidence. To ensure appropriate comparison, all metrics are computed using only residue centers (C$_\beta$ atoms) and ligand heavy atoms—matching the coarse-grained representation of distogram-based models, even for Boltz-1 which generates all-atom structures. Rigid alignment between predicted and ground truth structures assigns equal weight to ligand atoms and protein residues.

We report two complementary metrics: (1) ligand RMSD $<2\text{\AA}$ success rate, which evaluates global pose accuracy across the entire ligand within a $15\text{\AA}$ pocket context, and (2) a combined success rate requiring both ligand RMSD $<2\text{\AA}$ and LDDT-PLI $>0.8$. The latter is a stricter metric: LDDT-PLI is computed using a $6\text{\AA}$ radius, capturing local distance accuracy at the immediate protein-ligand interface rather than the broader pocket geometry.

\subsubsection{Per-Benchmark Performance}

Figure~\ref{fig:structure_all_benchmarks} presents structure prediction results across all benchmarks: FoldBench, PoseBusters, Runs N' Poses, and Proprietary crystal structures.

Despite operating without a diffusion module, TerraBind achieves competitive or superior ligand pose accuracy across nearly all benchmarks. On the ligand RMSD $<2\text{\AA}$ metric, TerraBind matches or exceeds Boltz-1, demonstrating that coarse-grained optimization from distance logits recovers poses comparable to diffusion-based approaches.

\begin{figure}[H]
    \centering
    \begin{subfigure}[b]{0.67\textwidth}
        \centering
        \includegraphics[width=\linewidth]{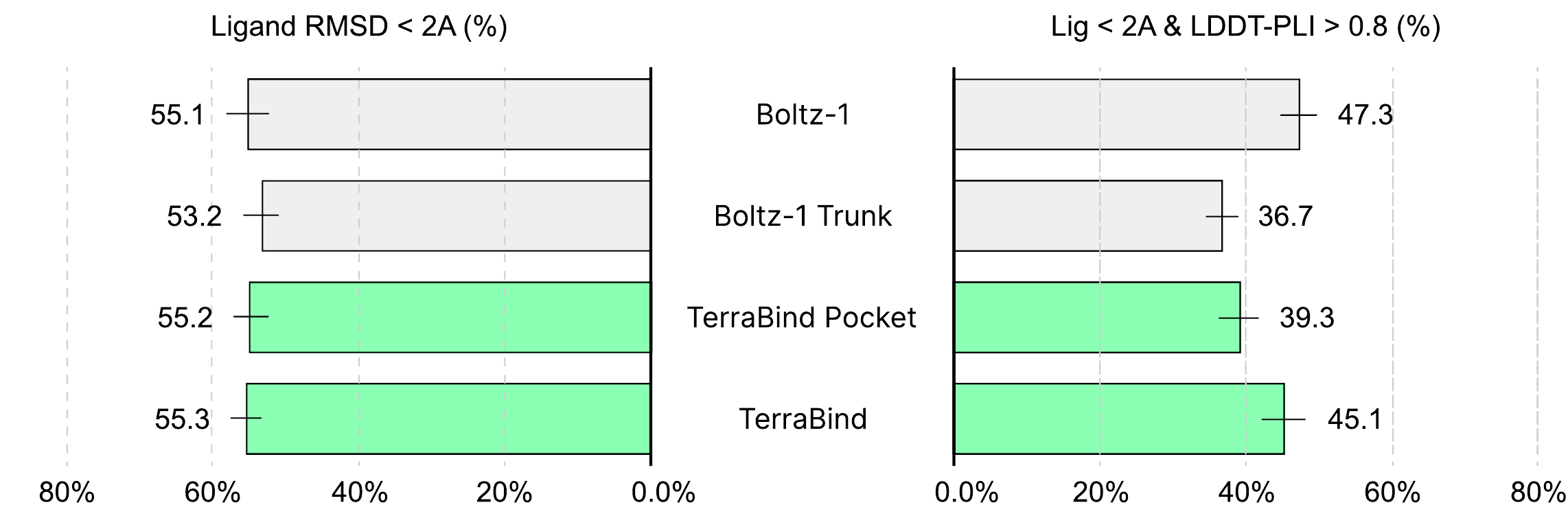}
        \caption{FoldBench ($n=556$)}
    \end{subfigure}
    \caption{\textbf{Structure prediction performance across benchmarks.} Each panel shows ligand RMSD $<2\text{\AA}$ success rate (left bars) and combined success rate requiring both RMSD $<2\text{\AA}$ and LDDT-PLI $>0.8$ (right bars). (a) FoldBench, a low-homology benchmark.}
    \label{fig:structure_all_benchmarks}
\end{figure}

\begin{figure}[H]\ContinuedFloat
    \centering
    \begin{subfigure}[b]{0.67\textwidth}
        \centering
        \includegraphics[width=\linewidth]{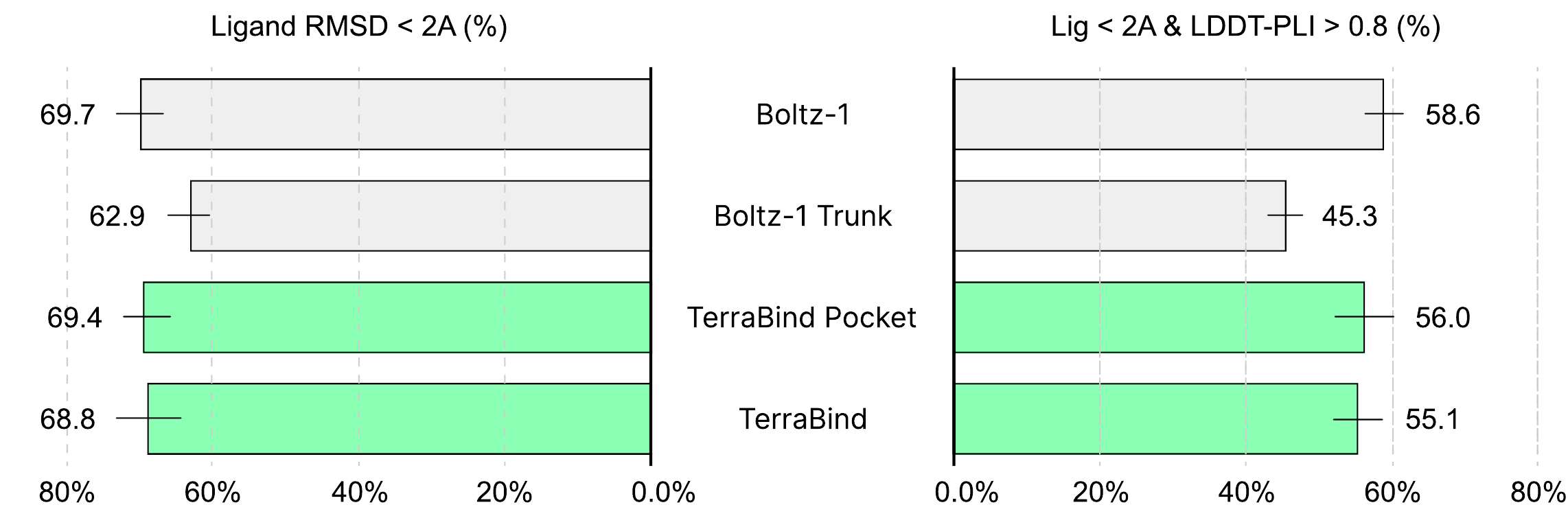}
        \caption{PoseBusters ($n=307$)}
    \end{subfigure}
    
    \vspace{0.4cm}
    
    \begin{subfigure}[b]{0.67\textwidth}
        \centering
        \includegraphics[width=\linewidth]{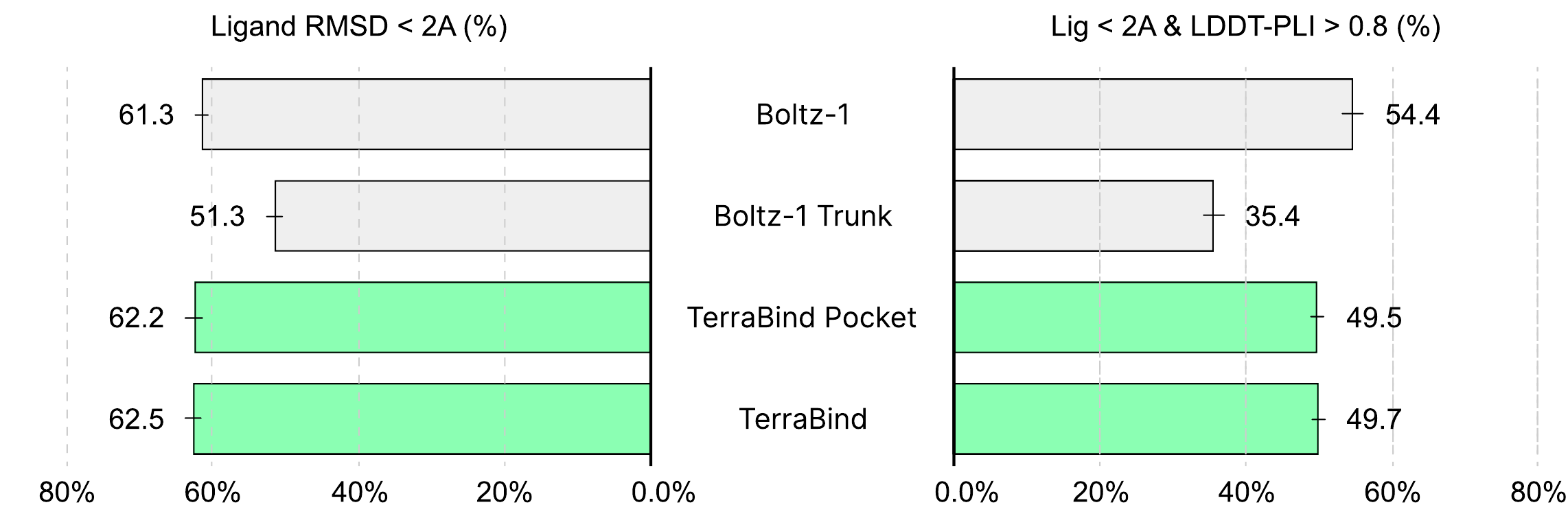}
        \caption{Runs N' Poses ($n=2{,}687$)}
    \end{subfigure}
    
    \vspace{0.4cm}
    
    \begin{subfigure}[b]{0.67\textwidth}
        \centering
        \includegraphics[width=\linewidth]{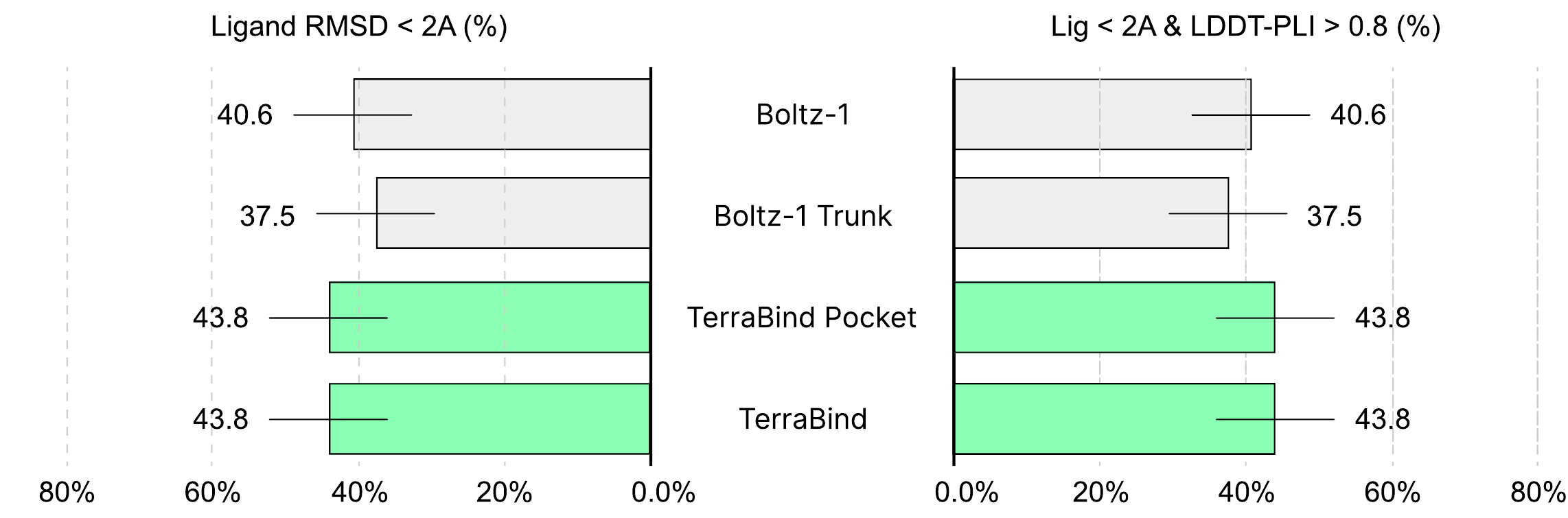}
        \caption{Proprietary crystal structures}
    \end{subfigure}
    \caption{\textbf{Structure prediction performance across benchmarks (cont.).} (b) PoseBusters, drug-like complexes released after 2021; (c) Runs N' Poses, high-resolution systems; (d) Proprietary crystal structures from active drug discovery programs.}
\end{figure}

    
    
    

A key finding is that TerraBind's pairformer learns substantially richer geometric representations than Boltz-1 Trunk. Applying identical coordinate optimization to both models (TerraBind vs. Boltz-1 Trunk), TerraBind consistently outperforms by a significant margin. This gap is especially pronounced in the combined success metric (RMSD $<2\text{\AA}$ and LDDT-PLI $>0.8$), reflecting TerraBind's superior accuracy at the immediate binding interface. TerraBind's strong performance on this stringent metric confirms that our training protocol yields richer structural features, which we believe also underlie the improved binding affinity predictions (Section~\ref{sec:affinity_results}).

TerraBind Pocket, operating on a compact 196-token context centered on the binding site, maintains robust performance despite the reduced context. This is unsurprising given that the model was trained with 256-token ligand-centered crops in Stages 2 and 3, closely matching the inference regime. The strong TerraBind Pocket results confirm that binding-relevant information is concentrated in the pocket region and full-protein context is unnecessary for accurate pose prediction.

\subsubsection{Distogram Entropy as Structural Confidence}

\begin{figure}[H]
    \centering
    \begin{subfigure}[b]{0.48\textwidth}
        \centering
        \includegraphics[width=\linewidth]{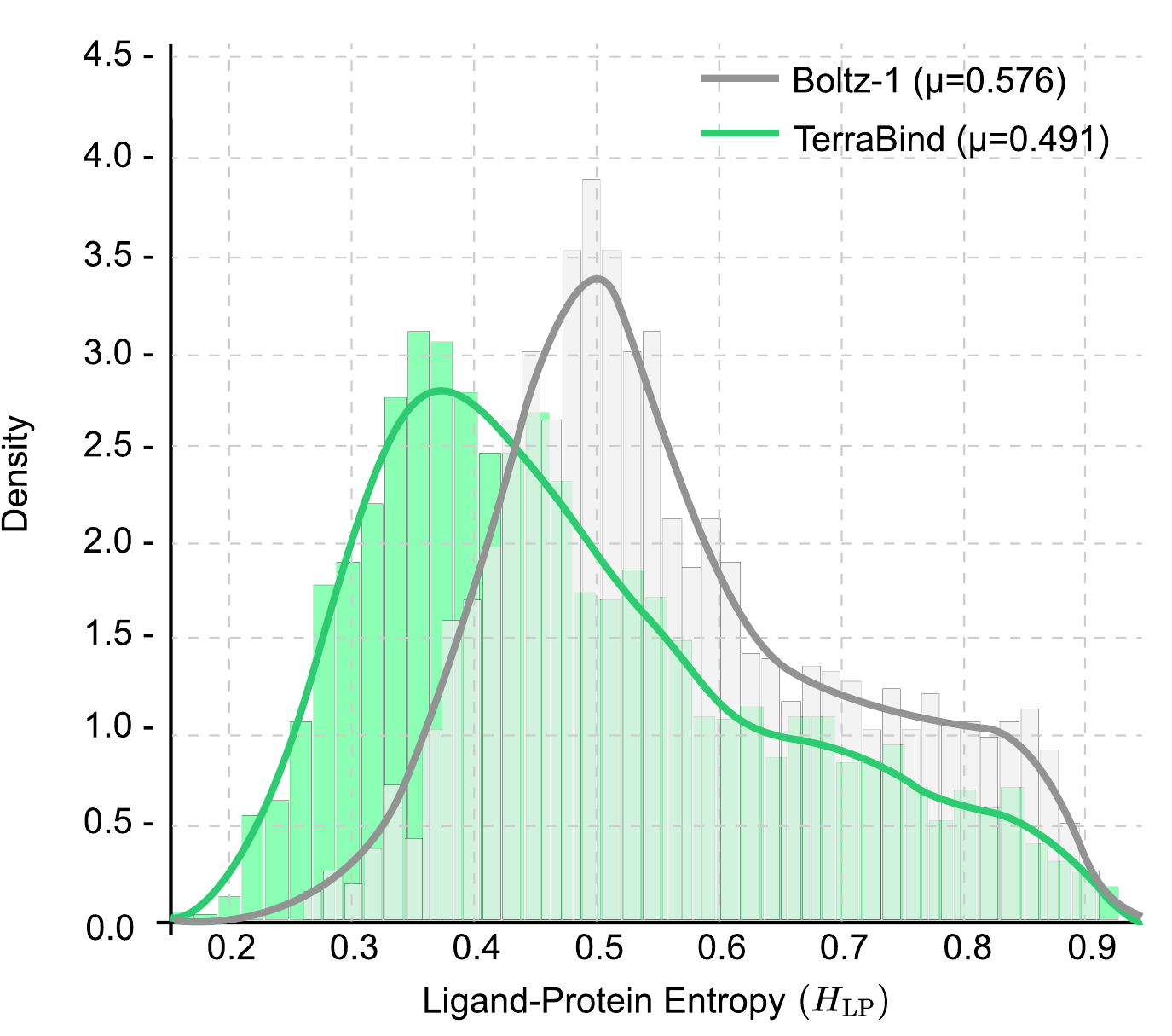}
        \caption{Ligand-protein entropy distributions.}
        \label{fig:entropy_comparison_bar}
    \end{subfigure}
    \hfill
    \begin{subfigure}[b]{0.48\textwidth}
        \centering
        \includegraphics[width=\linewidth]{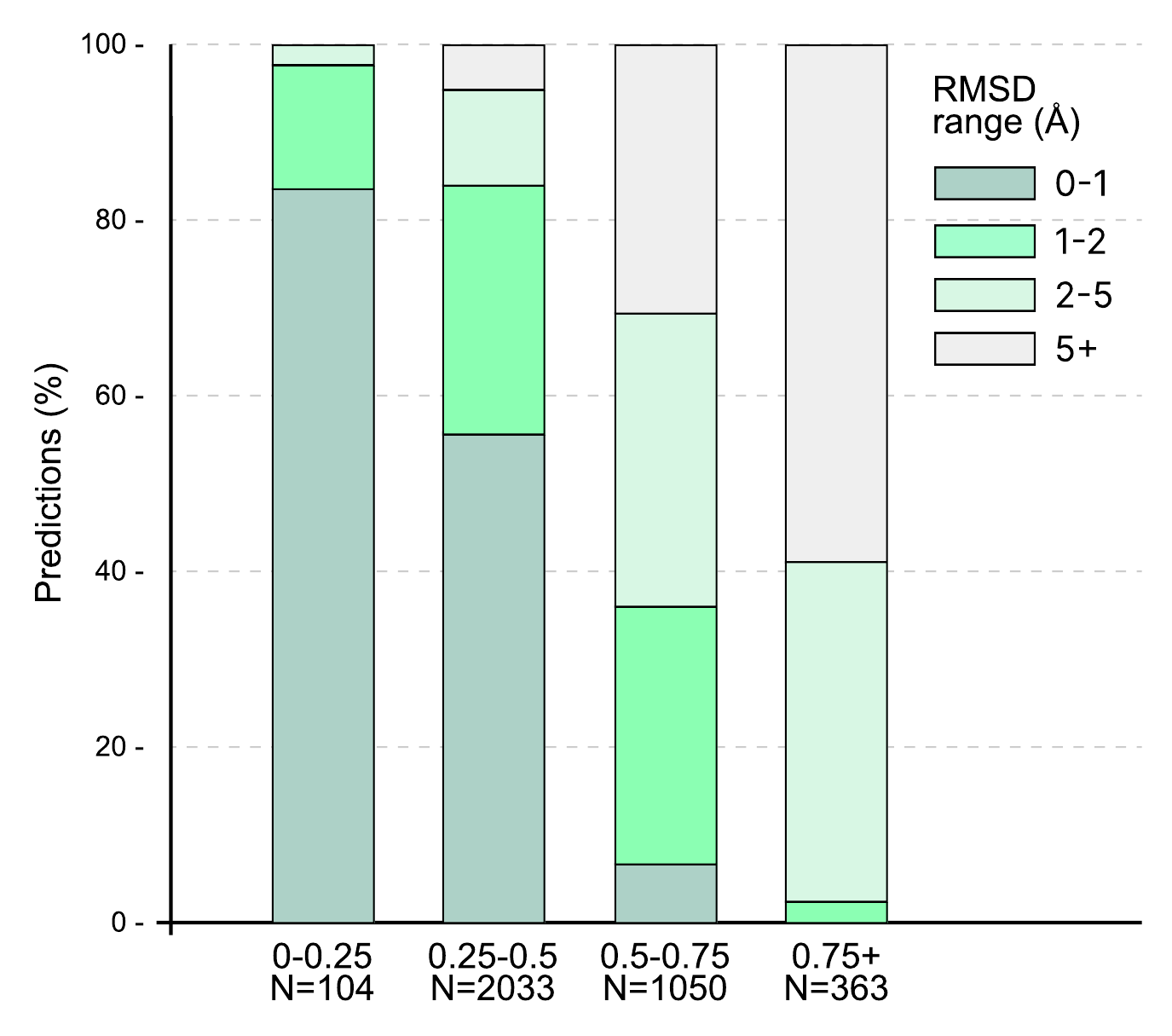}
        \caption{Ligand-protein entropy range vs.\ pose accuracy.}
        \label{fig:entropy_calibration}
    \end{subfigure}
    \caption{\textbf{Distogram entropy as structural confidence.} 
    (a) Distribution of ligand-protein entropy ($H_{\text{LP}}$) for TerraBind and Boltz-1 Trunk, aggregated across FoldBench, PoseBusters, and Runs N' Poses benchmarks. Lower entropy indicates more confident predictions. TerraBind's distribution is shifted toward lower values, achieving 15\% lower mean entropy than Boltz-1 Trunk.
    (b) Relationship between TerraBind predicted entropy and pose accuracy. Lower entropy predictions achieve substantially higher success rates across all RMSD thresholds, validating entropy as a built-in confidence metric.}
    \label{fig:entropy_analysis}
\end{figure}

Beyond pose accuracy, we examine whether the pairformer's predicted distance distributions provide meaningful uncertainty estimates. Figure~\ref{fig:entropy_analysis} shows two complementary analyses of ligand-protein entropy ($H_{\text{LP}}$).

Figure~\ref{fig:entropy_comparison_bar} compares the distribution of ligand-protein entropy between TerraBind and Boltz-1 Trunk, aggregated across FoldBench, PoseBusters, and Runs N' Poses. TerraBind's entropy distribution is shifted toward lower values, indicating substantially more confident distance predictions at the binding interface. TerraBind achieves mean $H_{\text{LP}} = 0.491$ compared to $0.576$ for Boltz-1 Trunk—a 15\% reduction in entropy.

Critically, this entropy correlates strongly with pose accuracy (Figure~\ref{fig:entropy_calibration}). Predictions with low entropy ($H_{\text{LP}} < 0.25$, $n=104$) achieve high success rates across all RMSD thresholds, while high-entropy predictions ($H_{\text{LP}} > 0.75$, $n=363$) rarely produce accurate poses. The intermediate bins ($H_{\text{LP}} \in [0.25, 0.5]$, $n=2{,}033$ and $H_{\text{LP}} \in [0.5, 0.75]$, $n=1{,}050$) show a monotonic decrease in accuracy with increasing entropy. This calibration emerges naturally from structure learning without explicit supervision, validating $H_{\text{LP}}$ as a model-intrinsic confidence metric that requires no separate confidence module.

\subsubsection{Inference Throughput}

TerraBind Pocket generates 10 pose samples in 1.045 seconds per complex compared to 27.8 seconds for Boltz-2 on equivalent hardware (single A6000 GPU, 196 tokens)---a 26.6$\times$ speedup that enables practical deployment for high-throughput binding affinity prediction.


We observe that TerraBind does not significantly benefit from generating more than 5 samples, whereas Boltz-1 continues to improve with additional samples and symmetry correction. This suggests that our optimization-based sampling produces less diverse poses than diffusion-based generation—an area for future improvement.

\subsection{Binding Affinity Prediction}
\label{sec:affinity_results}

Having established that TerraBind produces accurate structural representations efficiently, we now evaluate its binding affinity prediction capabilities. We assess performance across CASP16 and proprietary assay data, examine the calibration of our uncertainty quantification module, and demonstrate the value of structural fine-tuning for improving affinity predictions.

\subsubsection{Binding Affinity Correlation Performance}

A central design decision in the affinity module is eliminating the need for all-atom 3D structures as input. Our hypothesis is that the pair representations from the structural pairformer are primarily responsible for providing the affinity signal, making all-atom structure refinement unnecessary—at least for small molecule binding. Rather than conditioning on discretized pairwise distograms derived from costly all-atom 3D structure generation (typically via diffusion), we condition directly on the pairwise distogram bin probabilities that follow from the structural pairformer representations. This approach requires neither diffusion nor the coordinate optimization module described in Section~\ref{sec:pose_module}.

\begin{figure}[H]
    \centering
    \begin{subfigure}[b]{\textwidth}
        \centering
        \includegraphics[width=0.75\linewidth]{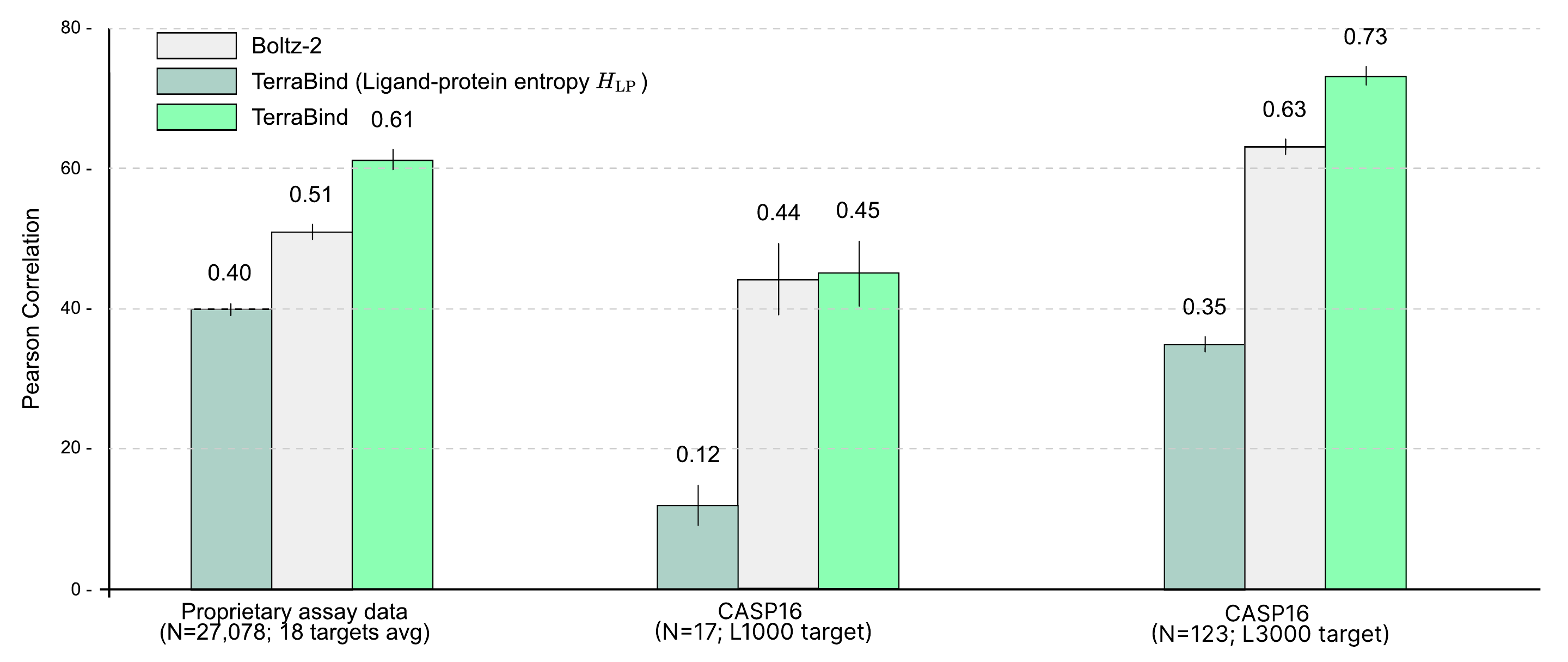}
        \caption{Pearson correlation across benchmark datasets.}
    \end{subfigure}
    
    \vspace{0.4cm}
    
    \begin{subfigure}[b]{\textwidth}
        \centering
        \includegraphics[width=1.0\linewidth]{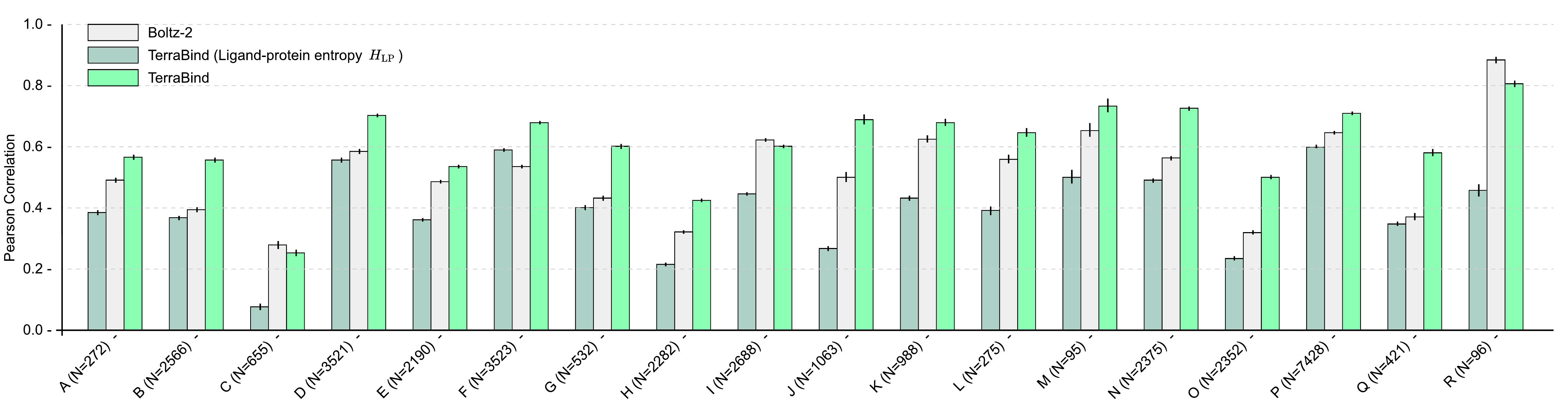}
        \caption{Per-target correlation comparison across proprietary assay data.}
    \label{fig:affinity_per_target}
    \end{subfigure}
    
    \caption{\textbf{Binding affinity prediction performance.} (a) Pearson correlation across benchmark datasets. On proprietary assay data, TerraBind outperforms Boltz-2 on 15 of 18 assayed targets. The zero-shot ligand-protein entropy ($H_{\text{LP}}$) also shows meaningful correlation with binding affinity. (b) Per-assay correlation comparison across all proprietary targets, demonstrating consistent improvements over Boltz-2 across diverse protein families.}
    \label{fig:affinity_overall}
\end{figure}

The structure prediction results (Section~\ref{sec:structure_performance}) demonstrate that these distogram bin probabilities are sufficient for generating high-accuracy ligand poses. Notably, we also find that the entropy derived from predicted bin probabilities correlates with binding affinity even in a zero-shot setting (Figure~\ref{fig:affinity_overall}). These observations motivated TerraBind's architecture, which forgoes all-atom 3D structure input entirely. Importantly, this enables substantially faster end-to-end affinity prediction inference.

Combined with COATI-3 global ligand encoding, this approach yields a model that exceeds Boltz-2 prediction accuracy. Figure~\ref{fig:affinity_overall} shows similar or improved Pearson correlation across all affinity benchmarks: both CASP16 targets and proprietary assay data covering 18 distinct protein targets from internal drug discovery campaigns. The proprietary data provides a particularly important evaluation, as it is more extensive, more diverse (spanning multiple chemical series), and uncurated—whereas published data is often curated and biased toward highly active compounds. Critically, TerraBind was trained exclusively on public data, making the proprietary benchmark a true test of out-of-distribution generalization.

\subsubsection{Structural Fine-Tuning Improves Affinity Predictions}

To demonstrate the value of structural information for binding affinity prediction, we designed an experiment to isolate the effect of structural fine-tuning. We performed a brief fine-tuning of the structural pairformer on a small set of proprietary crystallographic data, then evaluated the affinity module—without any retraining—using the fine-tuned structural representations as input.

\begin{figure}[H]
    \centering
    \includegraphics[width=0.48\linewidth]{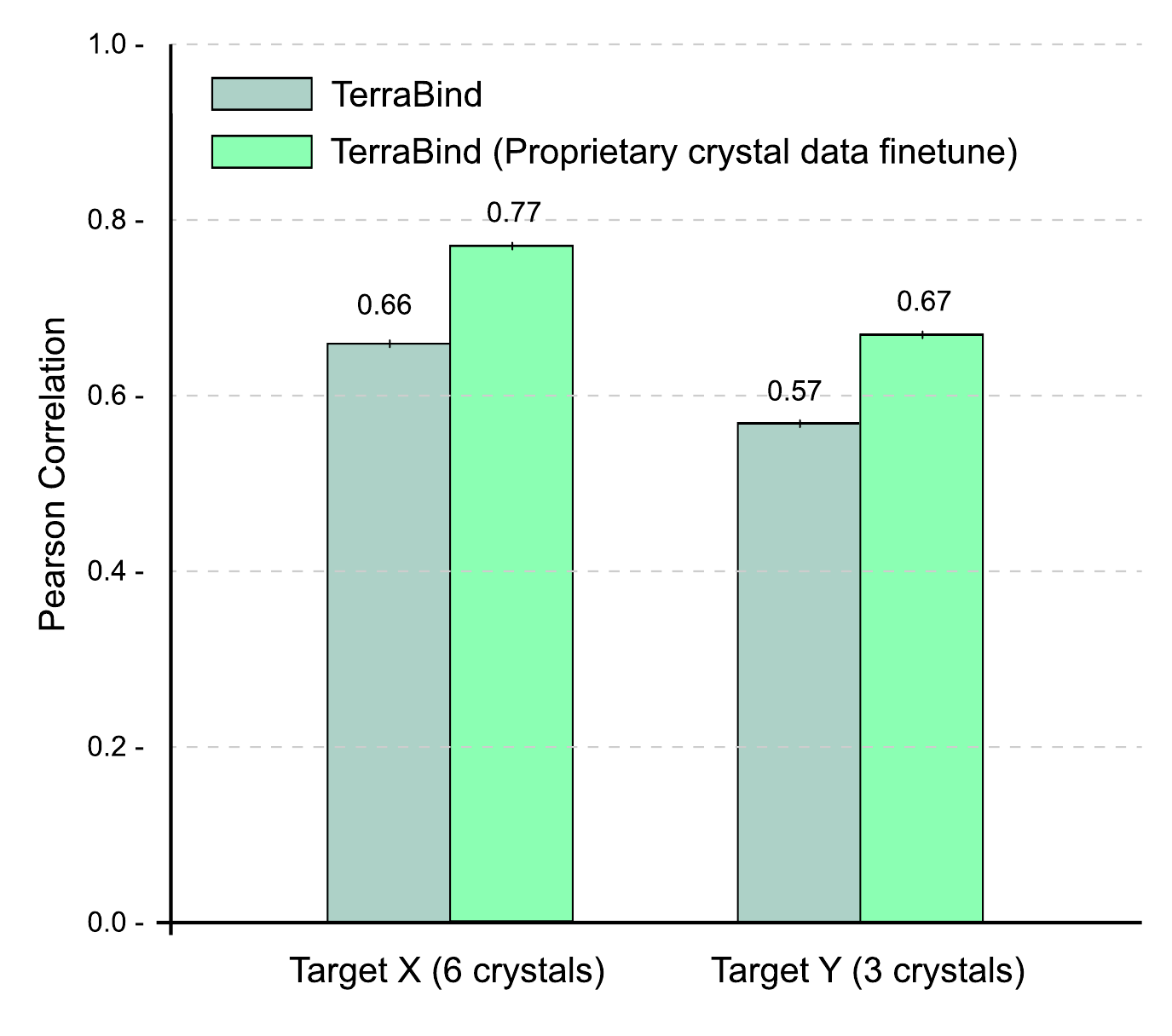}
    \caption{\textbf{Impact of structural fine-tuning on affinity prediction.} Pearson correlation for two unrelated targets ($<20\%$ sequence similarity) using the same TerraBind affinity module with different structural pairformer inputs: the base model versus a structurally fine-tuned variant trained on Proprietary crystal structures. The fine-tuning set contained 6 crystals for Target X and 3 crystals for Target Y.}
    \label{fig:internal_crystal_finetune}
\end{figure}

Figure~\ref{fig:internal_crystal_finetune} shows results for two targets present in the fine-tuning crystallographic data. Despite their low sequence similarity ($<20\%$), both targets show notable improvement in prediction correlation after structural fine-tuning. This demonstrates that even with very limited structural data and minimal training, we can improve affinity accuracy on thousands of unseen test molecules—without fine-tuning the affinity module itself. These results underscore the tight coupling between structural representation quality and downstream affinity prediction performance, and highlight the practical value of incorporating program-specific structural data when available.

\subsubsection{Binding Affinity Uncertainty Quantification}
\label{sec:epinet_marginals_calibration}

Beyond prediction accuracy, practical drug discovery requires well-calibrated uncertainty estimates for binding affinity predictions. Due to activity cliffs—where seemingly subtle molecular changes produce dramatic affinity differences—reliable error quantification is especially challenging and often neglected in general binding models~\cite{passaro2025boltz}. To our knowledge, we present the first uncertainty quantification module for general co-folded binding affinity models.

\begin{figure}[H]
    \centering
    \includegraphics[width=0.65\linewidth]{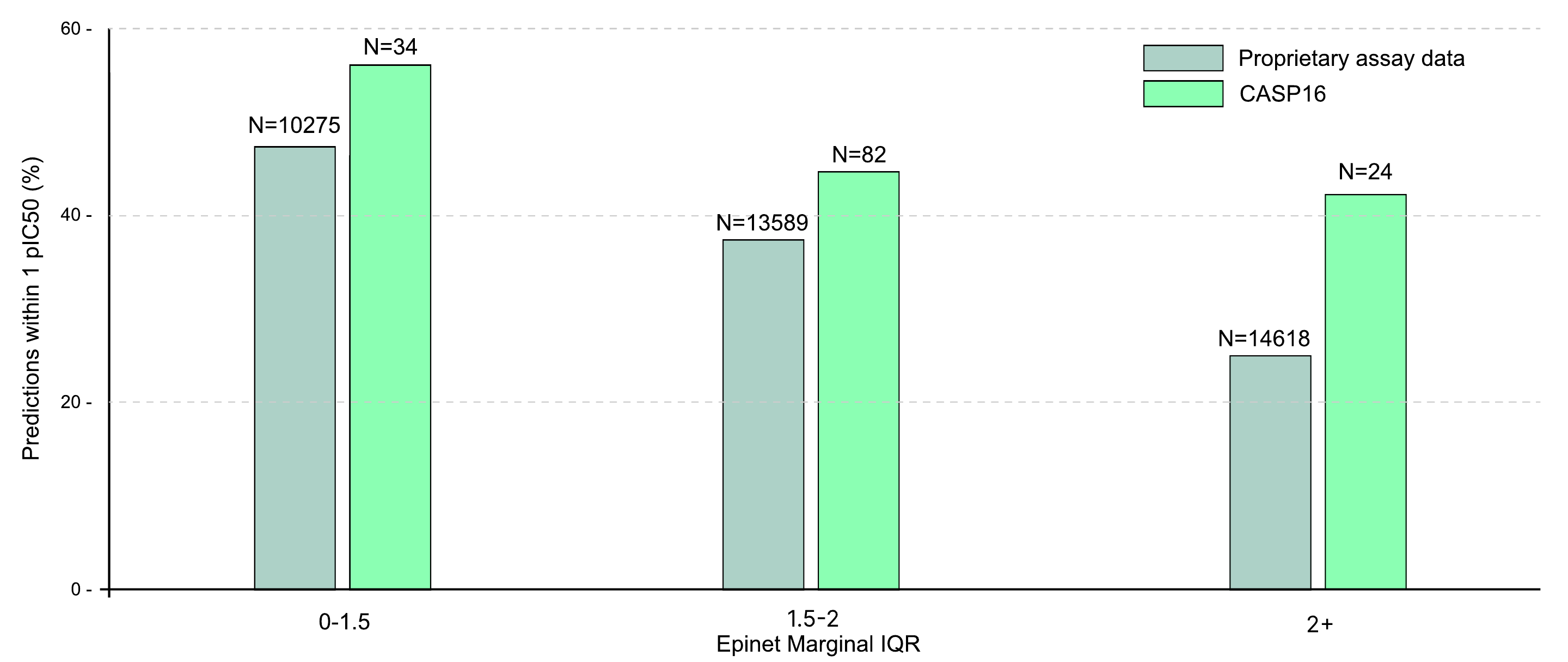}
    \caption{\textbf{Uncertainty calibration of the epinet module.} Success percentage (predictions within 1 pIC50 of true value) stratified by predicted epinet interquartile range (IQR). Higher IQR indicates greater model uncertainty. Lower uncertainty predictions achieve notably higher success rates, demonstrating well-calibrated confidence estimates across CASP16 and proprietary assay data.}
    \label{fig:epinet_marginals_calibration}
\end{figure}

To illustrate uncertainty calibration, Figure~\ref{fig:epinet_marginals_calibration} shows the relationship between the interquartile range (IQR) of epinet affinity prediction samples and the corresponding success rate—defined as the fraction of predictions falling within 1~pIC50 (one log unit) of the true value. Predictions with lower IQR (lower uncertainty) achieve notably higher success rates, a relationship that holds across a diverse set of protein targets: both CASP16 and proprietary assay data.

\subsubsection{Simulated DMTA Cycles with Continual Learning}

The results presented in Section~\ref{sec:epinet_marginals_calibration} reflect marginal epistemic uncertainty for individual binding affinity predictions. 
Additionally, one can construct joint predictive distributions by sharing epistemic indices across a batch of complexes as described in Section~\ref{sec:affinity_likelihood_module}.
In recent work~\cite{wang2025pretrained}, we demonstrate that this joint distribution approach enables superior molecule selection over multiple Design-Make-Test-Analyze (DMTA) cycles compared to traditional greedy selection strategies used in drug discovery. Additionally, we can also leverage a \textit{continual learning} scheme to quickly incorporate real-world observations into the model predictions without explicit retraining (Section~\ref{sec:continual_learning_details}).

\begin{figure}[H]
    \centering
    \includegraphics[width=0.58\linewidth]{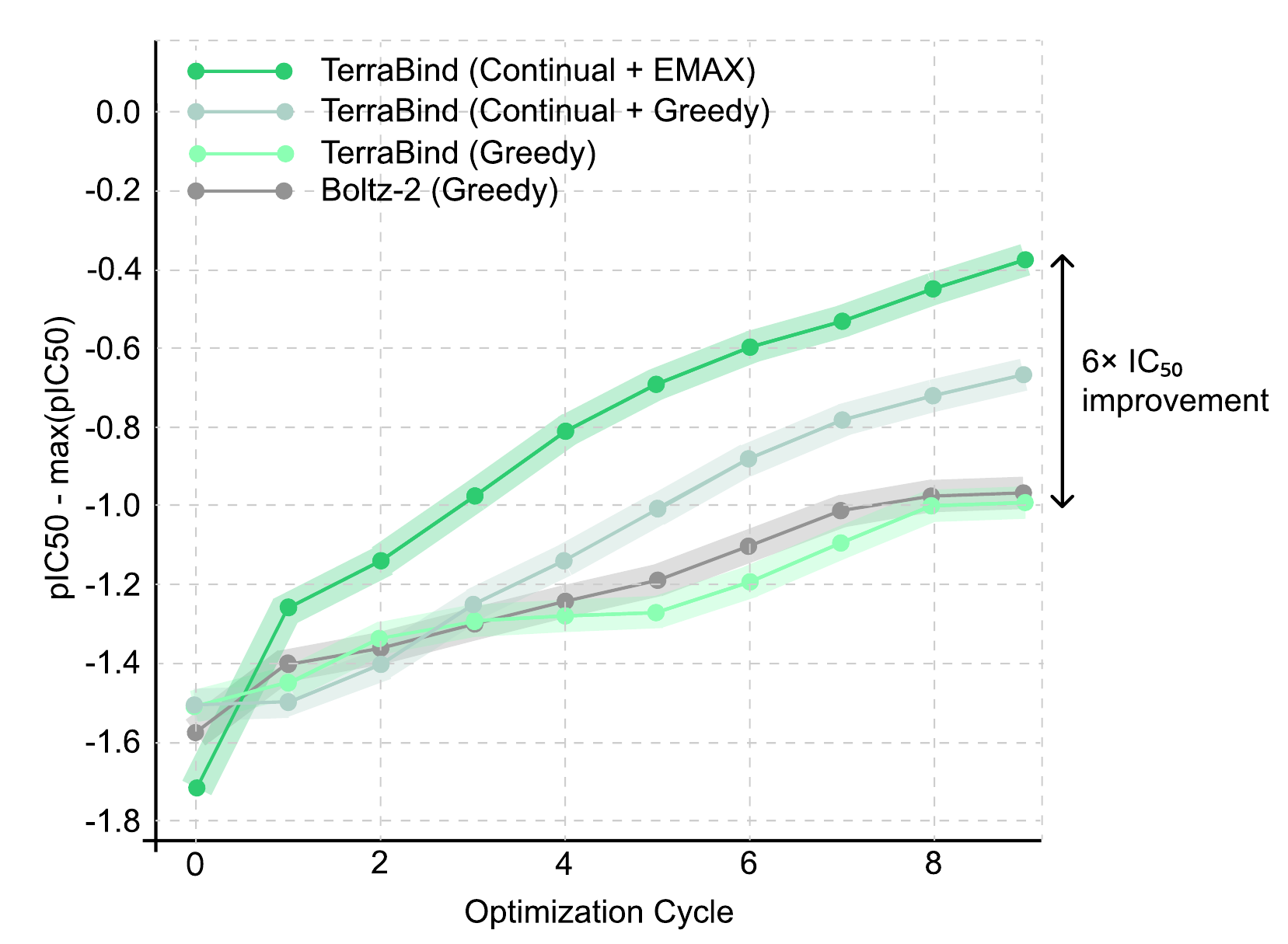}
    \caption{\textbf{Continual Learning Accelerates Hit Optimization.} Simulated DMTA cycle experiment using only $5$ selected compounds per cycle. The plot monitors the cumulative difference between the (predicted) max selected pIC50 compound and the max known pIC50 compound. We compare four different strategies: non-continual greedy-based approaches (Boltz2-Greedy and TerraBind-Greedy), a continual learning greedy-based approach (Continual TerraBind-Greedy), and a continual learning hedged selection approach (Continual TerraBind-EMAX). In the last cycle, we find a 6$\times$ IC50 affinity improvement with Continual TerraBind-EMAX versus the non-continual greedy approaches (note the y-axis here is in log-scale).}
    \label{fig:continual_dmta_figure}
\end{figure}

To demonstrate this DMTA selection enhancement, in Fig.~\ref{fig:continual_dmta_figure} we run a case study of a simulated DMTA cycle experiment using held-out proprietary data for Target I, where TerraBind and Boltz2 have seemingly comparable accuracy (see Fig.~\ref{fig:affinity_per_target}).
We consider a fixed selection pool consisting of over $2500$ small molecules with experimental affinity observations accumulated from internal drug campaign programs. This pool of molecules is especially challenging to optimize as it consists of numerous harsh activity cliffs, where slight molecule modifications can cause massive shifts in binding affinity. 
In the first DMTA cycle, the entire molecule pool is available for selection, and a selection strategy is used to obtain $5$ compounds for assay readout. The selected compounds are then removed from the available pool, and the process is repeated. In Fig.~\ref{fig:continual_dmta_figure}, we track the difference between the highest pIC50 in the entire pool of molecules, and the highest pIC50 observed from all molecules selected up to the current cycle. We compare four different selection strategies:
\textit{Boltz2-Greedy} is a greedy top-$5$ selection using the Boltz2 affinity model, \textit{TerraBind-Greedy} is a greedy top-$5$ selection using the TerraBind affinity model, \textit{Continual TerraBind-Greedy} is a greedy top-$5$ selection using the TerraBind affinity model where predictions have been updated at each cycle based on past cycle observations, and \textit{Continual TerraBind-EMAX} utilizes predicted joint distributions and the EMAX acquisition function (Eq.~\ref{eq:emax}) with batch size $5$ to make selections. Strategies that use continual learning use the Terrabind epinet module to quickly update joint predictions with observations obtained each cycle (Section~\ref{sec:continual_learning_details}). 

We find that while greedy approaches that utilize the base models (Boltz2-Greedy and TerraBind-Greedy) have similar poor performance for this target, the continual learning approaches quickly outperform. In particular, the EMAX acquisition function further outperforms greedy-based approaches, as EMAX can suitably hedge between compounds selected within a batch by penalizing inferred correlations between them. Note that neither the continual learning scheme nor the EMAX acquisition scheme can be readily applied using the Boltz2 model. They are only enabled via the epinet-based affinity likelihood module presented in this work. 


\section{Limitations}

\textbf{Coarse-grained approach.} Our generated structures do not possess all-atom resolution. While this can be a limitation for the downstream use of poses in computational chemistry pipelines~\cite{boltzabfe2025}, we demonstrate throughout this work that all-atom resolution is not necessary for accurate ML-driven predictions in practical small-molecule drug discovery. We also note that the simple optimization scheme outlined in Section~\ref{sec:pose_module} is not suitable for large-scale structure generation. In practice, we find that in larger systems, predicted pairwise distograms can be incompatible with a corresponding 3D point cloud, causing the optimization to struggle with convergence. Therefore, we suggest this method is best limited to small-molecule binding site contexts.

\textbf{Out-of-distribution structure prediction.} While poor generalization to unseen protein targets is a well-known deficiency of structural models~\cite{xu2025foldbench}, there is also a pervasive failure to address diverse regions of ligand chemical space, even when protein sequences are well-represented. This is evident in the predicted ligand-protein interface entropy, where known strong binders often exhibit high entropy. Such a conflict with experimental data suggests a chemical space deficiency in the structural models. Recently, it has been shown that the extensive expansion of ligand chemical space using synthetic data carefully generated by physics-based approaches can improve structural performance~\cite{dobles2025pearl}. We suggest that this could also have a significant impact on downstream binding affinity performance.

\textbf{Binding affinity context.} A fundamental limitation of binding affinity predictions from these models is their limited context, as they consider only the protein and molecule of interest. Measurements from cell-based assays or activity assays with complex mechanisms of action may be especially unsuitable for modeling using a restricted protein-ligand binding context. Since much public experimental data is derived from such assays, this introduces significant noise and artifacts during training, despite strategies to minimize these effects. Future work focusing on careful curation, categorization of assay data, and models with assay-conditioned predictions may make better use of this data.

\textbf{Uncertainty quantification.} A major limitation in the current binding affinity likelihood module is that the epinet architecture yields Gaussian predicted marginal distributions~\cite{osband2022fine}. This is problematic, as the distributions tend to be optimistic; they are typically centered around the TerraBind affinity prediction with vanishing density for lower binding affinities, suggesting an unrealistically low false-positive likelihood. In practice, models in certain domains can have high false-positive rates that are not well-calibrated with the predicted likelihood. In recent work~\cite{wang2025pretrained}, we find that pretraining the prior network using a reference process to generate synthetic datasets with non-Gaussian marginals—mimicking the bounded and skewed distributions of real assays in practical drug discovery—can better mitigate this effect induced by naive prior networks.

\section{Conclusion}

In this work, we introduced TerraBind, a multimodal foundation model designed to resolve the computational bottlenecks that currently limit the application of high-accuracy cofolding models to large-scale virtual screening of small molecules. By re-evaluating the architectural necessities of structure-based drug design, we demonstrated that expensive, all-atom diffusion-based structure generation is not a prerequisite for accurate pose generation or binding affinity prediction.

Our approach leverages frozen pretrained encoders—specifically COATI-3 for ligand representation~\cite{kaufman2024coati,kaufman2024latent} and ESM-2 for protein sequences~\cite{lin2023evolutionary}—coupled with a streamlined pairformer trunk. This architecture allows TerraBind to learn rich, coarse-grained representations that directly inform binding affinity. The results highlight several critical advancements:

\textbf{Efficient Diffusion-Free Pose Generation.} Despite relying on a coarse-grained optimization strategy rather than a generative diffusion model, TerraBind achieves ligand pose accuracy competitive with state-of-the-art baselines on benchmarks such as FoldBench, PoseBusters, and Runs N' Poses.

\textbf{State-of-the-Art Affinity Prediction.} By leveraging these rich structural representations and a pretrained ligand encoder, TerraBind achieves superior binding affinity prediction accuracy on both public benchmarks (CASP16~\cite{zhang2025assessment}) and proprietary assay data compared to existing models like Boltz-2. Combined, the end-to-end pipeline delivers a 26-fold increase in throughput over existing models.

\textbf{Data-Efficient Fine-Tuning.} We demonstrated that the model's modular design allows for effective structural fine-tuning on limited proprietary crystallographic data, significantly improving binding affinity performance on these targets.

\textbf{Calibrated Confidence Estimation.} The introduction of a dedicated affinity likelihood module provides robust uncertainty quantification. We showed that the model's predicted error ranges correlate well with empirical success rates, addressing a long-standing need for reliable confidence metrics in virtual screening.

\textbf{Enhanced Hit Optimization via Epinet-Based Likelihoods.} The proposed epinet-based likelihood module enables both continual learning and the EMAX acquisition function, facilitating a more effectively hedged selection approach that outperforms traditional greedy strategies in simulated drug discovery DMTA cycles.

Ultimately, TerraBind establishes a new paradigm for structure-based drug discovery, proving that coarse-grained structural insights are sufficient for highly accurate binding affinity and structure prediction. This framework makes it much more feasible to deploy high-fidelity deep learning predictions across practical, library-scale drug discovery campaigns, and importantly for training iterations, it becomes feasible to leverage billion-scale proprietary datasets delivered from Terray's EMMI platform.

\section*{Acknowledgments}
We thank the Terray Therapeutics team for their support and contributions to this work, NVIDIA for GPU efficiency expertise and cuEquivariance modules, and the broader scientific community for creating the open-source tools and datasets that enabled this research.


\clearpage
\appendix

\section{Implementation Details}

\subsection{Pairformer Architecture}
\label{sec:pairformer_arch}

The pairformer trunk follows the architecture introduced in Ref.~\cite{abramson2024accurate}, with key modifications for computational efficiency. Each pairformer layer updates pair representations through two core operations: (1) \textit{triangle attention}, and (2) \textit{triangle multiplication}.

\textbf{Triangle attention} operates on pair representations by attending along rows or columns with biases derived from edges that complete triangles:
\begin{equation}
z'_{ij} = z_{ij} + \sum_k \text{softmax}_k\left( \frac{q_{ij} \cdot k_{\alpha}}{\sqrt{d_h}} + b_{\beta} \right) v_{\alpha}
\quad \text{where} \quad
(\alpha, \beta) = 
\begin{cases}
(ik, jk) & \text{starting node} \\
(kj, ki) & \text{ending node}
\end{cases}
\label{eq:triangle_attention}
\end{equation}
where $q_{ij}, k_{\alpha}, v_{\alpha}$ are linear projections of the pair representations, $d_h$ is the head dimension, and $b_{\beta}$ is a learned bias derived from $z_{\beta}$—the edge completing the triangle $(i, j, k)$. The starting node variant attends along rows (fixed $i$), while the ending node variant attends along columns (fixed $j$).

\textbf{Triangle multiplication} propagates information through multiplicative gating over edges sharing a common node:
\begin{equation}
z'_{ij} = z_{ij} + \text{Linear}\left( \sum_k g_a(z_{\gamma}) \odot g_b(z_{\delta}) \right)
\quad \text{where} \quad
(\gamma, \delta) = 
\begin{cases}
(ik, jk) & \text{outgoing} \\
(ki, kj) & \text{incoming}
\end{cases}
\label{eq:triangle_mult}
\end{equation}
where $g_a(\cdot)$ and $g_b(\cdot)$ are distinct gated linear projections and $\odot$ denotes element-wise multiplication. These operations encode geometric consistency: if residues $i$ and $j$ are both close to residue $k$, then $i$ and $j$ should be close to each other.

We use a 48-layer pairformer with 4-head attention and 128-dimensional pair representations. Unlike Ref.~\cite{abramson2024accurate}, we omit single sequence representations entirely—these are only required for the downstream diffusion module, which is absent in TerraBind. This reduces the structure module parameter count from $\sim$147M to $\sim$27M.

We implement triangle attention and triangle multiplication using optimized CUDA kernels from NVIDIA cuEquivariance (v0.6.0)~\cite{cuequivariance}, with all operations performed in bfloat16 precision. This provides $1.6\times$ training speedup and $3\times$ inference speedup (depending on context length) compared to standard implementations while maintaining numerical stability. For full architectural details, we refer readers to Ref.~\cite{abramson2024accurate}.

\subsection{Distance Binning and Structure Learning}
\label{sec:structure_module_details}

\subsubsection{Distance Bin Configuration}

Following Ref.~\cite{wohlwend2025boltz}, we project the final pair representations into $N_{\text{bins}} = 64$ pairwise distance bins. The bins are configured as follows: 62 bins evenly spaced from $2\text{\AA}$ to $22\text{\AA}$ (bin width $\approx 0.32\text{\AA}$), plus two boundary bins for covalent-range distances ($<2\text{\AA}$) and long-range interactions ($>22\text{\AA}$). Distances are predicted between all ligand heavy atoms and protein residue centers (represented by C$_\beta$ atoms, or C$_\alpha$ for glycine). The pairformer produces a distogram of pairwise distance bin probability distributions, $p(d_{ij}^b)$, where $i,j$ index atom pairs and $b \in \{1, \ldots, N_{\text{bins}}\}$ enumerates distance bins.

\subsubsection{Structure Learning Loss}
\label{sec:structure_loss}

The distogram is trained via categorical cross-entropy over distance bins. For each pair of atoms $(i,j)$, we compute the pointwise loss:
\begin{equation}
\ell_{ij} = -\sum_{b=1}^{N_{\text{bins}}} y_{ij}^b \log p(d_{ij}^b)
\end{equation}
where $y_{ij}^b$ is the one-hot encoding of the ground truth distance bin and $p(d_{ij}^b) = \text{softmax}(\mathbf{z}_{ij})_b$ is the predicted probability for bin $b$.

To emphasize binding-relevant geometry, we apply pair-type-specific weights. Let $\mathcal{M}$ denote the set of valid (non-padding, non-diagonal) atom pairs. We construct a weight matrix $W$ where each entry $w_{ij}$ depends on the pair type:
\begin{equation}
w_{ij} = 
\begin{cases}
w_{\text{LL}} & \text{if } i,j \in \text{ligand} \\
w_{\text{LP}} & \text{if } (i \in \text{ligand}, j \in \text{protein}) \text{ or } (i \in \text{protein}, j \in \text{ligand}) \\
w_{\text{PP}} & \text{if } i,j \in \text{protein}
\end{cases}
\end{equation}

The total structure loss is then computed as:
\begin{equation}
\mathcal{L}_{\text{struct}} = \frac{\sum_{(i,j) \in \mathcal{M}} w_{ij} \cdot \ell_{ij}}{\sum_{(i,j) \in \mathcal{M}} w_{ij}}
\label{eq:structure_loss}
\end{equation}

Pair types are distinguished as follows:
\begin{itemize}
    \item \textbf{Ligand-ligand (LL)}: pairs where both atoms belong to ligand molecules, including both intra-ligand pairs (within the same molecule) and inter-ligand pairs (between different ligand molecules in multi-ligand systems)
    \item \textbf{Ligand-protein (LP)}: pairs between ligand atoms and protein residue centers
    \item \textbf{Protein-protein (PP)}: pairs where both atoms are protein residue centers, including both intra-chain pairs (within the same protein chain) and inter-chain pairs (between different chains in multi-chain complexes)
\end{itemize}

The weights $w_{\text{LL}}$, $w_{\text{LP}}$, and $w_{\text{PP}}$ are adjusted across training stages to progressively emphasize binding interface geometry (see Table~\ref{tab:training_stages}). In Stage 1, all pair types are weighted equally ($w_{\text{LL}} = w_{\text{LP}} = w_{\text{PP}} = 1$). In Stage 2, we upweight ligand-ligand ($w_{\text{LL}} = 2$) and ligand-protein ($w_{\text{LP}} = 5$) pairs to focus learning on the binding site. Stage 3 returns to equal weights while training exclusively on experimental structures.

\subsubsection{Expected Pairwise Distance}
\label{sec:expected_distance}

From the predicted bin distributions, we compute expected pairwise distances as a probability-weighted average of bin centers:
\begin{equation}
\hat{d}_{ij} = \sum_{b=1}^{N_{\text{bins}}} p(d_{ij}^b) \cdot c_b
\label{eq:expected_pair_distance}
\end{equation}
where $c_b$ is the midpoint of bin $b$. For boundary bins, we use $c_1 = 1.5\text{\AA}$ (covalent) and $c_{64} = 24.5\text{\AA}$ (long-range). These expected distances define the binding pocket:
\begin{equation}
\mathcal{P} = \{j \in \text{protein} : \exists\, i \in \text{ligand},\, \hat{d}_{ij} < 15\text{\AA}\}
\label{eq:pocket_definition}
\end{equation}
This pocket definition serves as input context for both the affinity module (Section~\ref{sec:affinity_module}) and the pose module (Section~\ref{sec:pose_module}).


\subsubsection{Normalized Pairwise Entropy}
\label{sec:pairwise_entropy}

We compute a normalized (information-theoretic) entropy for each pairwise distance distribution to quantify prediction confidence:
\begin{equation}
H(d_{ij}) = -\frac{1}{\log N_{\text{bins}}} \sum_{b=1}^{N_{\text{bins}}} p(d_{ij}^b) \log p(d_{ij}^b)
\label{eq:pairwise_entropy}
\end{equation}
Normalization by $\log N_{\text{bins}}$ ensures $H(d_{ij}) \in [0, 1]$, where $H \approx 0$ indicates a peaked distribution (confident prediction) and $H \approx 1$ indicates a uniform distribution (uncertain prediction).

\subsubsection{Aggregated Entropy Metrics}

We aggregate pairwise entropies by pair type to obtain summary confidence metrics. Let $L$ denote the set of ligand atoms and $\mathcal{P}$ the binding pocket residues (Eq.~\ref{eq:pocket_definition}).

\textbf{Ligand-ligand entropy} averages over all ligand atom pairs:
\begin{equation}
H_{\text{LL}} = \frac{1}{|L|(|L| - 1)} \sum_{\substack{i,j \in L \\ i \neq j}} H(d_{ij})
\label{eq:ll_entropy}
\end{equation}

\textbf{Ligand-protein entropy} averages over ligand-pocket interactions:
\begin{equation}
H_{\text{LP}} = \frac{1}{|L| \cdot |\mathcal{P}|} \sum_{i \in L} \sum_{j \in \mathcal{P}} H(d_{ij})
\label{eq:lp_entropy}
\end{equation}

\textbf{Protein-protein entropy} averages over pocket residue pairs:
\begin{equation}
H_{\text{PP}} = \frac{1}{|\mathcal{P}|(|\mathcal{P}| - 1)} \sum_{\substack{i,j \in \mathcal{P} \\ i \neq j}} H(d_{ij})
\label{eq:pp_entropy}
\end{equation}

Note that $H_{\text{LL}}$ is computed over all ligand atoms, while $H_{\text{PP}}$ is restricted to pocket residues within the $15\text{\AA}$ cutoff. The diagonal terms ($i = j$) are excluded from $H_{\text{LL}}$ and $H_{\text{PP}}$ as self-distances are trivially zero.

The ligand-protein entropy $H_{\text{LP}}$ serves as the primary confidence metric for structure prediction and provides a zero-shot binding affinity signal: lower $H_{\text{LP}}$ correlates with both higher pose accuracy and stronger binding (Section~\ref{sec:affinity_results}).

\subsection{Multi-Stage Training Protocol}

We employ an effective batch size of 128 distributed across 4 NVIDIA H100 nodes and a 48-layer pairformer with 4-multihead pairwise attention. The TerraBind model was trained for a total of 105k steps following the multi-stage protocol described in Table~\ref{tab:training_stages}.

\begin{table}[h]
\centering
\small
\caption{Multi-stage structure prediction training protocol. Data columns show sampling probabilities; loss weight columns show relative weighting for each pair type.}
\label{tab:training_stages}
\begin{tabular}{c c c c c c c c c c}
\toprule
& & & \multicolumn{3}{c}{\textbf{Data Sampling}} & \multicolumn{3}{c}{\textbf{Loss Weights}} \\
\cmidrule(lr){4-6} \cmidrule(lr){7-9}
\textbf{Stage} & \textbf{Steps} & \textbf{Tokens} & PDB & AFDB & BindingDB & LL & LP & PP \\
\midrule
1 & 70k & 384 & 0.45 & 0.25 & 0.30 & 1$\times$ & 1$\times$ & 1$\times$ \\
2 & 20k & 256 & 0.50 & 0.00 & 0.50 & 2$\times$ & 5$\times$ & 1$\times$ \\
3 & 15k & 256 & 1.00 & 0.00 & 0.00 & 1$\times$ & 1$\times$ & 1$\times$ \\
\bottomrule
\end{tabular}
\end{table}

\subsection{Pocket Cropping Algorithm}
\label{sec:pocket_crop_algorithm}

The pocket cropping algorithm (Algorithm~\ref{alg:pocket_crop}) enables efficient inference on large proteins by restricting computation to the binding site region. The algorithm operates in two stages: (1) an initial full-protein pairformer pass to identify pocket residues, and (2) a refined pass on the cropped context.

In the initial pass, pocket residues are identified using a $22\text{\AA}$ cutoff on the expected pairwise distances $\hat{d}_{ij}$ (Eq.~\ref{eq:expected_pair_distance}):
\begin{equation}
\mathcal{P}_{\text{init}} = \{j \in \text{protein} : \exists\, i \in \text{ligand},\, \hat{d}_{ij} < 22\text{\AA}\}
\label{eq:pocket_init}
\end{equation}
This larger cutoff (compared to the $15\text{\AA}$ used for affinity prediction) ensures sufficient context for the refined pass.

The algorithm prioritizes including all ligand atoms and all pocket residues. If the combined token count exceeds the budget (rare), pocket residues are truncated by distance to ligand, keeping those closest. If the token budget permits additional context, the algorithm performs contiguous sequence expansion: it identifies clusters of pocket residues within each protein chain (merging residues separated by $\leq 3$ positions), then adds neighboring residues in order of sequence proximity to these clusters. This expansion strategy preserves local secondary structure context while respecting the computational budget.

\begin{algorithm}[h]
\caption{Pocket-Based Context Cropping}
\label{alg:pocket_crop}
\small
\begin{algorithmic}[1]
\STATE \textbf{Input:} Tokenized structure, budget $N_{\text{max}}$, pocket indices $\mathcal{P}_{\text{init}}$, expected distances $\{\hat{d}_{ij}\}$
\STATE \textbf{Output:} Cropped tokenized structure $\mathcal{C}$

\STATE // \textit{Step 1: Always include entire ligand}
\STATE $\mathcal{L} \leftarrow \{i : \text{token}_i.\text{type} = \text{LIGAND}\}$
\STATE $\mathcal{C} \leftarrow \mathcal{L}$

\STATE // \textit{Step 2: Include pocket residues (truncating if necessary)}
\STATE $\mathcal{P} \leftarrow \{i : \text{token}_i.\text{idx} \in \mathcal{P}_{\text{init}}\}$
\IF{$|\mathcal{L}| + |\mathcal{P}| > N_{\text{max}}$}
    \STATE $\text{budget}_{\text{pocket}} \leftarrow N_{\text{max}} - |\mathcal{L}|$
    \STATE Sort $j \in \mathcal{P}$ by expected distance to ligand $\min_{k \in \mathcal{L}} \hat{d}_{jk}$ (ascending)
    \STATE $\mathcal{P}_{\text{keep}} \leftarrow \text{first } \text{budget}_{\text{pocket}} \text{ tokens from sorted } \mathcal{P}$
    \STATE $\mathcal{C} \leftarrow \mathcal{C} \cup \mathcal{P}_{\text{keep}}$
\ELSE
    \STATE $\mathcal{C} \leftarrow \mathcal{C} \cup \mathcal{P}$
\ENDIF

\STATE // \textit{Step 3: Contiguous sequence expansion (if under budget)}
\IF{$|\mathcal{C}| < N_{\text{max}}$}
    \STATE $\text{candidates} \leftarrow []$
    \FOR{each protein chain $c$ intersecting $\mathcal{C}$}
        \STATE $\mathcal{S}_c \leftarrow \{ \text{res\_idx}(i) : i \in \mathcal{C} \cap \text{chain}_c \}$ \hfill $\triangleright$ pocket residues in chain
        \STATE Form clusters $\{(s_k, e_k)\}$ from $\mathcal{S}_c$ where residue gap $\le 3$
        \FOR{each residue $r$ in chain $c$ not in $\mathcal{C}$}
            \STATE $d_{\text{seq}} \leftarrow \min_{k} (\min(|r - s_k|, |r - e_k|))$ \hfill $\triangleright$ dist to cluster boundary
            \STATE $\text{candidates}.\text{append}((d_{\text{seq}}, r))$
        \ENDFOR
    \ENDFOR
    \STATE Sort $\text{candidates}$ by $d_{\text{seq}}$ (ascending)
    \FOR{$(d_{\text{seq}}, r) \in \text{candidates}$}
        \IF{$|\mathcal{C}| \geq N_{\text{max}}$} \STATE \textbf{break} \ENDIF
        \STATE $\mathcal{C} \leftarrow \mathcal{C} \cup \{ \text{token}(r) \}$
    \ENDFOR
\ENDIF

\STATE // \textit{Step 4: Finalize}
\STATE Filter structure to retain only tokens in $\mathcal{C}$ and induced bonds
\STATE \textbf{return} $\mathcal{C}$
\end{algorithmic}
\end{algorithm}

\subsection{Pose Module}
\label{sec:pose_module_details}

The pose module generates 3D coordinates from predicted distograms via gradient-based optimization (Algorithm~\ref{alg:optimization}). Given the pairformer's distance bin distributions, we first extract expected pairwise distances $\hat{d}_{ij}$ as probability-weighted averages of bin centers (Eq.~\ref{eq:expected_pair_distance}). These expected distances define a target distance matrix that we use to optimize 3D coordinates via gradient descent on the joint ligand-pocket system.

We define the binding pocket $\mathcal{P}$ as protein residues within $15\text{\AA}$ of any ligand atom based on predicted ligand-protein distances (Eq.~\ref{eq:pocket_definition}). For the joint optimization over $N = L + |\mathcal{P}|$ points (ligand atoms plus pocket residues), we minimize the mean squared distance error:
\begin{equation}
\mathcal{L}_{\text{opt}} = \frac{1}{N(N-1)} \sum_{\substack{i,j \\ i \neq j}} w_{ij} \cdot \left( \|\mathbf{x}_i - \mathbf{x}_j\|_2 - \hat{d}_{ij} \right)^2
\label{eq:optimization_loss}
\end{equation}
where $\mathbf{x}_i \in \mathbb{R}^3$ are the coordinates being optimized and $\hat{d}_{ij}$ are the expected distances from the pairformer. All pair types (ligand-ligand, ligand-protein, and protein-protein) are weighted equally ($w_{ij} = 1$).

Coordinates are initialized from a standard normal distribution $\mathbf{x}_i \sim \mathcal{N}(0, \mathbf{I}_3)$ and optimized using Adam with learning rate $\alpha = 1.0$. Optimization terminates upon convergence (loss change $< 10^{-3}$ for 20 consecutive iterations) or after a maximum of 5000 iterations; in practice, convergence typically occurs within $\sim$500 iterations. After optimization, we perform rigid alignment with ground truth structures using equal weights for ligand and protein atoms.

\begin{algorithm}[H]
\caption{Joint Ligand-Pocket Coordinate Optimization}
\label{alg:optimization}
\small
\begin{algorithmic}[1]
\STATE \textbf{Input:} Pairformer logits $\mathbf{Z}^{LL} \in \mathbb{R}^{L \times L \times B}$, $\mathbf{Z}^{PP} \in \mathbb{R}^{P \times P \times B}$, $\mathbf{Z}^{LP} \in \mathbb{R}^{L \times P \times B}$
\STATE \textbf{Input:} Bin centers $\{c_b\}_{b=1}^{B}$, pocket cutoff $d_{\text{cut}} = 15\text{\AA}$
\STATE \textbf{Output:} Optimized coords $\mathbf{X}_{\text{lig}}$, $\mathbf{X}_{\text{pocket}}$, and pocket indices $\mathcal{P}$
\STATE \hfill $\triangleright$ $L$ = ligand heavy atoms, $P$ = protein C$\beta$ atoms, $B$ = distance bins
\STATE
\STATE // \textit{Step 1: Compute expected distance matrices from logits}
\FOR{each pair type $t \in \{\text{LL}, \text{PP}, \text{LP}\}$}
    \STATE $p_{ij}^b \leftarrow \text{softmax}(\mathbf{Z}^t_{ij})_b$ \hfill $\triangleright$ bin probabilities
    \STATE $\hat{D}_{ij}^t \leftarrow \sum_{b=1}^{B} p_{ij}^b \cdot c_b$ \hfill $\triangleright$ expected distance (Eq.~\ref{eq:expected_pair_distance})
    \STATE $\hat{\mathbf{D}}^t \leftarrow [\hat{D}_{ij}^t]$ \hfill $\triangleright$ expected distance matrix
\ENDFOR
\STATE
\STATE // \textit{Step 2: Identify pocket residues}
\STATE $\mathcal{P} \leftarrow \{j \in [P] : \min_{i \in [L]} \hat{D}_{ij}^{LP} < d_{\text{cut}}\}$
\STATE $K \leftarrow |\mathcal{P}|, \quad N \leftarrow L + K$
\STATE
\STATE // \textit{Step 3: Build reference distance matrix}
\STATE $\mathbf{R} \leftarrow \mathbf{0}^{N \times N}$
\STATE $\mathbf{R}_{[1:L,\, 1:L]} \leftarrow \hat{\mathbf{D}}^{LL}$ \hfill $\triangleright$ ligand-ligand block
\STATE $\mathbf{R}_{[L+1:N,\, L+1:N]} \leftarrow \hat{\mathbf{D}}^{PP}[\mathcal{P}, \mathcal{P}]$ \hfill $\triangleright$ pocket-pocket block
\STATE $\mathbf{R}_{[1:L,\, L+1:N]} \leftarrow \hat{\mathbf{D}}^{LP}[:, \mathcal{P}]$ \hfill $\triangleright$ ligand-pocket block
\STATE $\mathbf{R}_{[L+1:N,\, 1:L]} \leftarrow \mathbf{R}_{[1:L,\, L+1:N]}^\top$ \hfill $\triangleright$ symmetry
\STATE
\STATE // \textit{Step 4: Optimize coordinates via gradient descent}
\STATE $\mathbf{X} \sim \mathcal{N}(\mathbf{0}, \mathbf{I})$, \quad $\mathbf{X} \in \mathbb{R}^{N \times 3}$
\STATE Initialize Adam optimizer with learning rate $\alpha = 1.0$
\STATE $\text{stable} \leftarrow 0$
\FOR{$t = 1$ to $T_{\text{max}} = 5000$}
    \STATE $D_{ij} \leftarrow \|\mathbf{x}_i - \mathbf{x}_j\|_2$ \hfill $\triangleright$ current pairwise distances
    \STATE $\mathcal{L} \leftarrow \frac{1}{N(N-1)} \sum_{i \neq j} (D_{ij} - R_{ij})^2$
    \STATE $\mathbf{X} \leftarrow \text{Adam\_step}(\mathbf{X}, \nabla_{\mathbf{X}} \mathcal{L})$
    \IF{$|\mathcal{L}^{(t)} - \mathcal{L}^{(t-1)}| < 10^{-3}$}
        \STATE $\text{stable} \leftarrow \text{stable} + 1$
        \IF{$\text{stable} \geq 20$}
            \STATE \textbf{break}
        \ENDIF
    \ELSE
        \STATE $\text{stable} \leftarrow 0$
    \ENDIF
\ENDFOR
\STATE
\STATE \textbf{return} $\mathbf{X}_{[1:L]}$, $\mathbf{X}_{[L+1:N]}$, $\mathcal{P}$
\end{algorithmic}
\end{algorithm}

\subsection{Continual learning scheme}
\label{sec:continual_learning_details}

Here we employ a pathwise conditioning approach rooted in Gaussian Process (GP) regression \cite{matheron1973intrinsic, wilson2020efficiently}. We treat the ensemble of predictions from the Epinet as samples from an implied prior stochastic process $f \sim \mathcal{GP}(\mu, K)$. Instead of defining an explicit kernel function, we utilize the empirical covariance of the ensemble outputs to approximate the kernel $K$.
Given a set of observed data points $\mathcal{D}_{obs} = \{(\mathbf{x}_i, y_i)\}_{i=1}^N$ and a set of unobserved query points $\mathcal{X}_{*}$, we update the ensemble predictions directly using the linear update rule for conditional Gaussian simulations. The updated prediction vector $\hat{\mathbf{y}}_{new}$ at the query points is computed as:

\begin{equation}
\hat{\mathbf{y}}_{new} = \hat{\mathbf{y}}_{prior} + \hat{K}_{*}(\hat{K} + \sigma_{obs}^2 I)^{-1} \left( \mathbf{y}_{true} - \hat{\mathbf{y}}_{obs} - \boldsymbol{\epsilon} \right)
\label{eq:pathwise_update}
\end{equation}
where $\hat{\mathbf{y}}_{prior}$ denotes the original Epinet predictions at $\mathcal{X}_{*}$, $\hat{\mathbf{y}}_{obs}$ denotes predictions at the observed locations, and $\mathbf{y}_{true}$ is the vector of ground truth observations. The matrices $\hat{K}$ and $\hat{K}_{*}$ represent the empirical covariance between observed points, and between unobserved and observed points, respectively. We also inject independent Gaussian noise $\boldsymbol{\epsilon} \sim \mathcal{N}(0, \sigma_{obs}^2 I)$, with $\sigma_{obs}=0.5$, into the residual term for each sample path \cite{wilson2020efficiently, garnelo2018conditional}. This simple heuristic allows for rapid Bayesian conditioning on new observations without the expense of retraining.

\subsection{Structural Fine-tuning}
\label{sec:structural_finetuning}

To fine-tune our structural model on Proprietary crystal structures, we use the standard structural model training protocol with the following modifications: a smaller learning rate ($10^{-5}$), shorter training duration (5k steps), and a low proprietary crystal sampling rate ($0.01\%$) relative to experimental PDB structures. We observe that more aggressive fine-tuning (higher learning rates or sampling rates) tends to yield models that overfit on internal data and generalize poorly on public structure benchmarks.

\subsection{Boltz Inference Details}

Unless otherwise specified, Boltz-1 and Boltz-2 inferences were performed with default inference parameters. For timing comparisons with TerraBind, we use the same NVIDIA cuEquivariance (v0.6.0) package~\cite{cuequivariance}, mixed precision (bfloat16), and hardware (single A6000 GPU) to ensure fair comparison.

\section{Results details}

\subsection{Runtime Decomposition and Efficiency Analysis}
\label{sec:runtime_analysis}

In Figure~\ref{fig:speedup}, we showed a detailed latency decomposition comparing TerraBind against Boltz-2, benchmarked on a single NVIDIA A6000 GPU for a protein-ligand complex of 196 tokens. Note that these runtime metrics exclude Multiple Sequence Alignment (MSA) generation. The reported latencies represent an end-to-end structure and potency inference, generating 10 pose samples and a binding affinity prediction. The observed $26.6\times$ aggregate speedup is primarily driven by the elimination of the iterative diffusion denoising process. As shown in the ``Pose'' component, Boltz-2 requires $25.92$ seconds for coordinate generation. In contrast, our batched pose optimization requires only $0.17$ seconds, representing a reduction of over two orders of magnitude. Furthermore, the ``Trunk'' latency is reduced from $1.53$ seconds (Boltz-2) to $0.87$ seconds (TerraBind). This $1.75\times$ acceleration stems from our streamlined architecture, which utilizes a 48-layer Pairformer (compared to 64 layers in Boltz-2) and eliminates the single sequence representation track. Notably, the reported TerraBind trunk latency includes the computational cost of the pre-trained encoders (ESM-2 for proteins and COATI-3 for ligands), yet still significantly outperforms the baseline. Additionally, unlike Boltz-2, TerraBind does not employ a separate confidence module; the auxiliary head for affinity prediction incurs negligible overhead ($0.06$ seconds), validating our design choice to perform potency tasks directly on latent structural embeddings without requiring full coordinate reconstruction. In a ``minimal affinity setup''---defined as TerraBind without pose generation versus Boltz-2 restricted to 5 diffusion samples---we observe a $17.01\times$ effective speedup for a 196 token context.





\subsection{PDB-Only Fine-Tuning (Stage 3)}

\begin{figure}[H]
    \centering
    \includegraphics[width=0.42\linewidth]{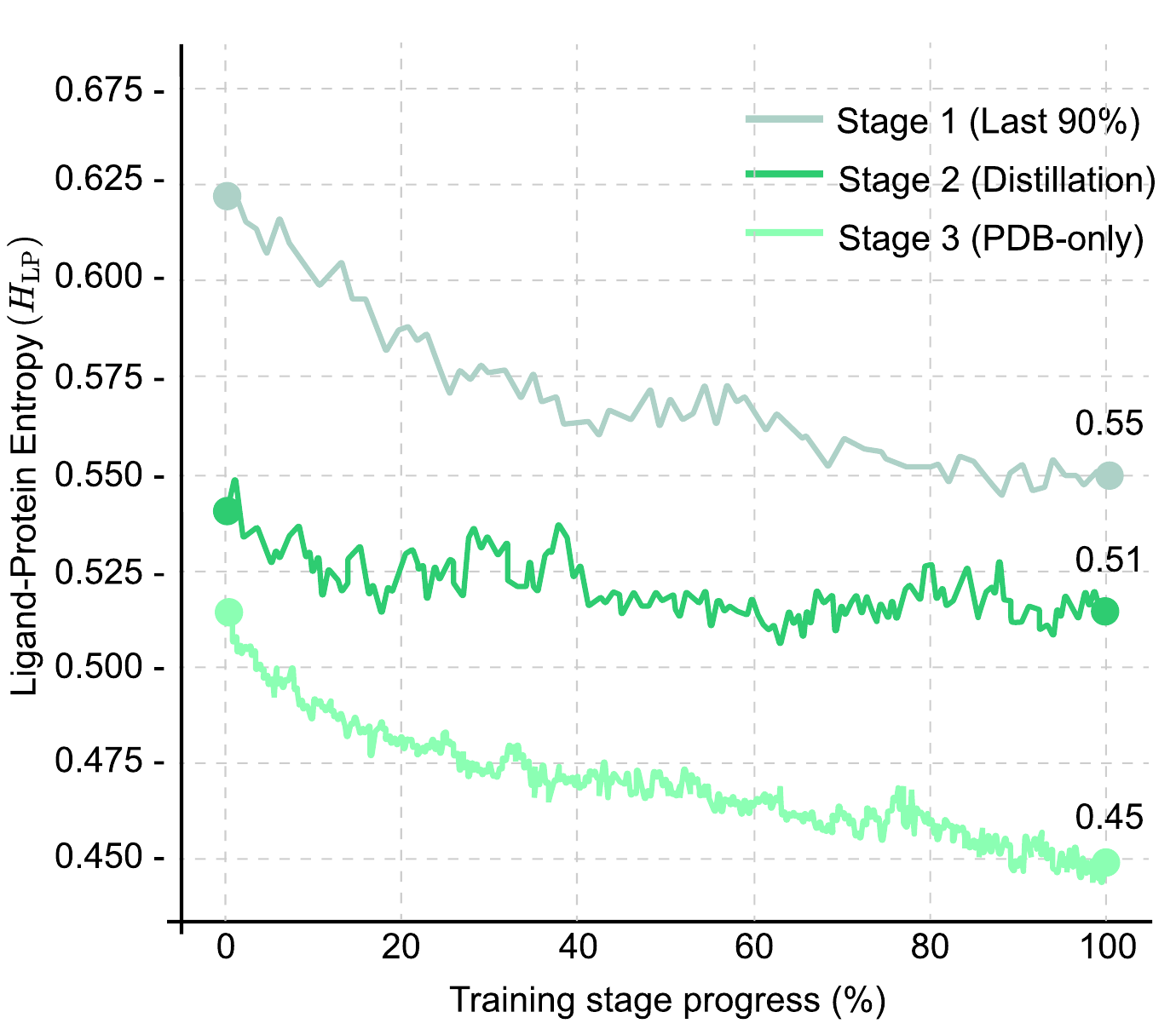}
    \caption{\textbf{Impact of Stage 3 PDB-only fine-tuning on distogram entropy.} Ligand-protein entropy over the three training stages. Stage 3 substantially reduces entropy, indicating more confident geometric predictions when trained on highest-quality experimental data.}
    \label{fig:stage3_entropy}
\end{figure}

Our multi-stage training protocol concludes with Stage 3: fine-tuning exclusively on experimental PDB structures with equal loss weights. Figure~\ref{fig:stage3_entropy} shows the impact of the three training stages on distogram confidence (held-out validation set). Stage 3 fine-tuning shifts the entropy distribution toward lower values, indicating the model becomes more confident about binding geometry when trained exclusively on high-quality experimental structures. This entropy reduction occurs without degrading RMSD performance, suggesting the model learns to place probability mass more decisively on the correct binding mode rather than hedging across alternatives. We hypothesize that distillation data (AFDB, BindingDB predictions) in earlier stages, while valuable for diversity, introduces some geometric ambiguity that is resolved by final fine-tuning on ground truth structures.

\subsection{Protein Encoder Ablations}

\begin{figure}[H]
    \centering
    \begin{subfigure}[b]{0.4\textwidth}
        \centering
        \includegraphics[width=\linewidth]{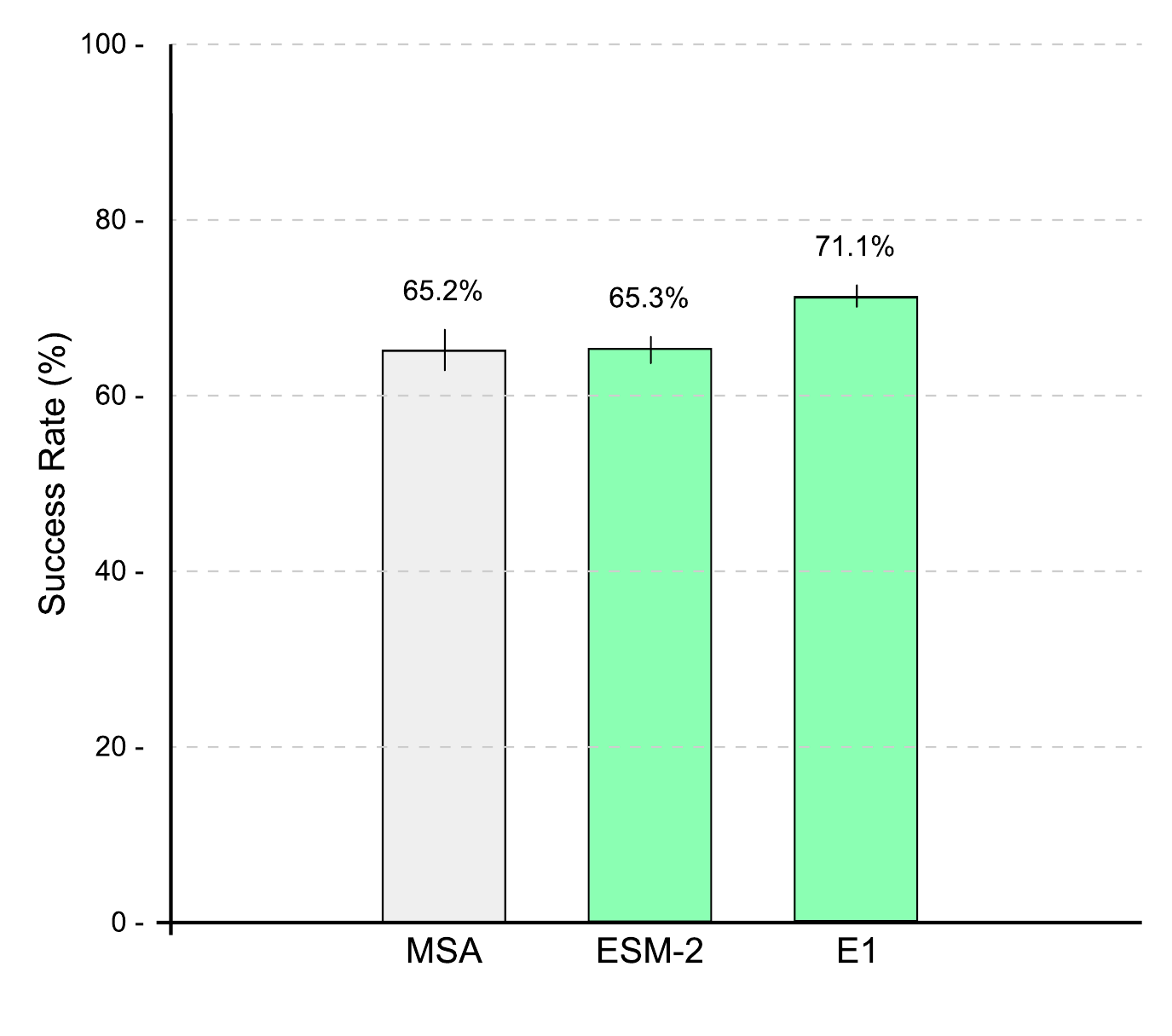}
        \caption{Protein RMSD $<2$\AA\ success rate.}
        \label{fig:encoder_ablation_rmsd}
    \end{subfigure}
    \begin{subfigure}[b]{0.4\textwidth}
        \centering
        \includegraphics[width=\linewidth]{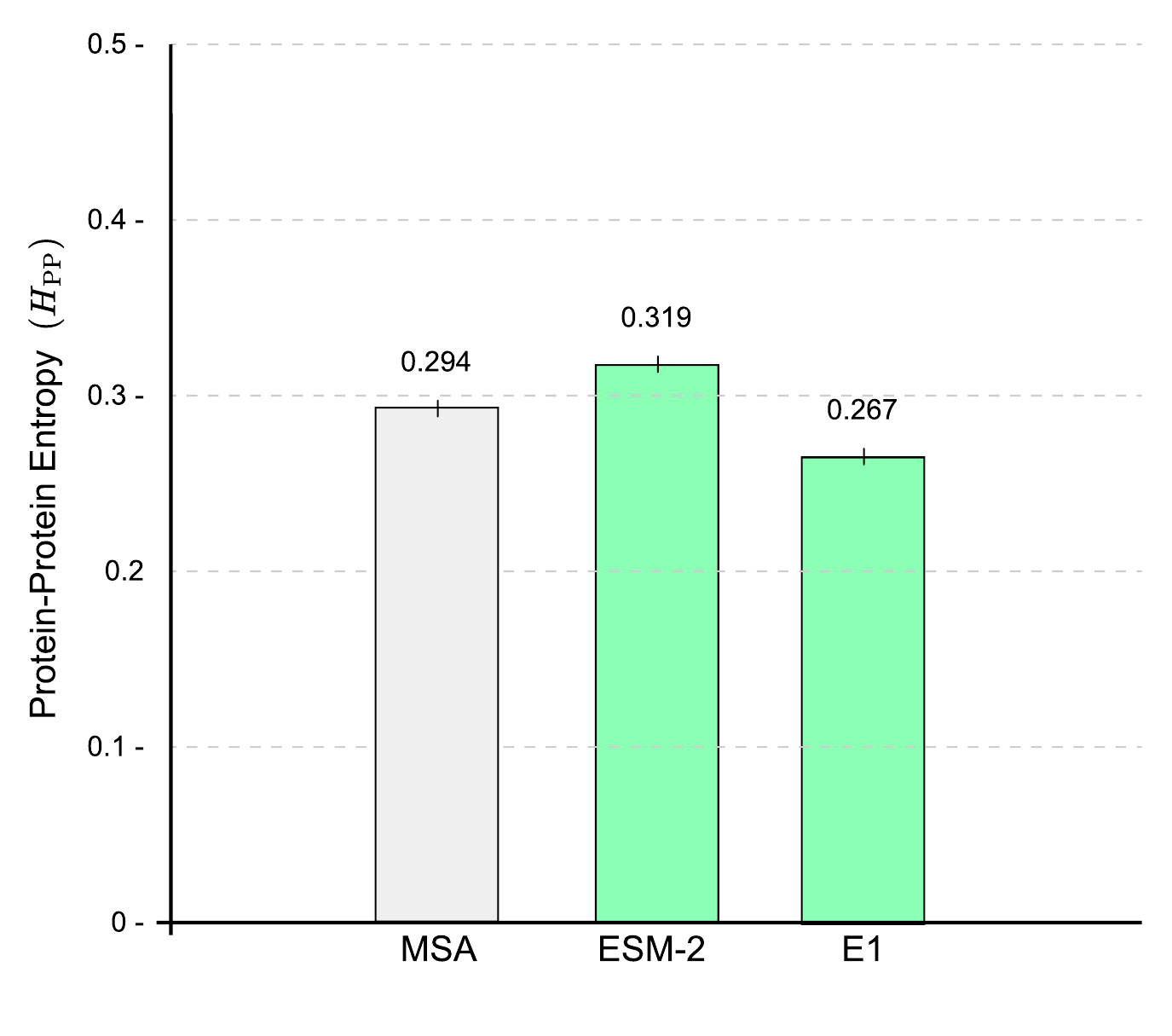}
        \caption{Protein-protein entropy $H_{\text{PP}}$.}
        \label{fig:encoder_ablation_entropy}
    \end{subfigure}
    \caption{\textbf{Protein encoder ablation.} 
    (a) Protein RMSD $<2$\AA\ success rate and (b) protein-protein entropy $H_{\text{PP}}$ aggregated across FoldBench, PoseBusters, and Runs N' Poses for TerraBind (ESM-2, 650M), TerraBind-E1 (Profluent-E1, 600M), and MSA via Boltz-1 Trunk. TerraBind-E1 achieves the highest protein RMSD success rate and the lowest $H_{\text{PP}}$, indicating more confident and accurate protein structure predictions.}
    \label{fig:entropy_analysis_ablations}
\end{figure}

To evaluate the impact of protein encoder choice on structure prediction, we trained model variants with different protein encoders using the same training regime as TerraBind. The ligand encoder (COATI-3 Allegro) remained fixed across experiments. We compare ESM-2 (650M), E1 (600M)~\cite{jain2025e1}, and MSA-based encoding. For the MSA baseline, we use the representations from Boltz-1 Trunk and apply our coordinate optimization for pose generation. Figure~\ref{fig:entropy_analysis_ablations} reports protein RMSD $<2\text{\AA}$ success rate and protein-protein entropy ($H_{PP}$) aggregated across FoldBench, PoseBusters, and Runs N' Poses benchmarks. Notably, E1 outperforms both ESM-2 and MSA-based encoding, achieving the highest success rate and lowest entropy.

\subsection{Limited Binding Site Context for Affinity Inference}

\begin{figure}[H]
    \centering
    \includegraphics[width=0.44\linewidth]{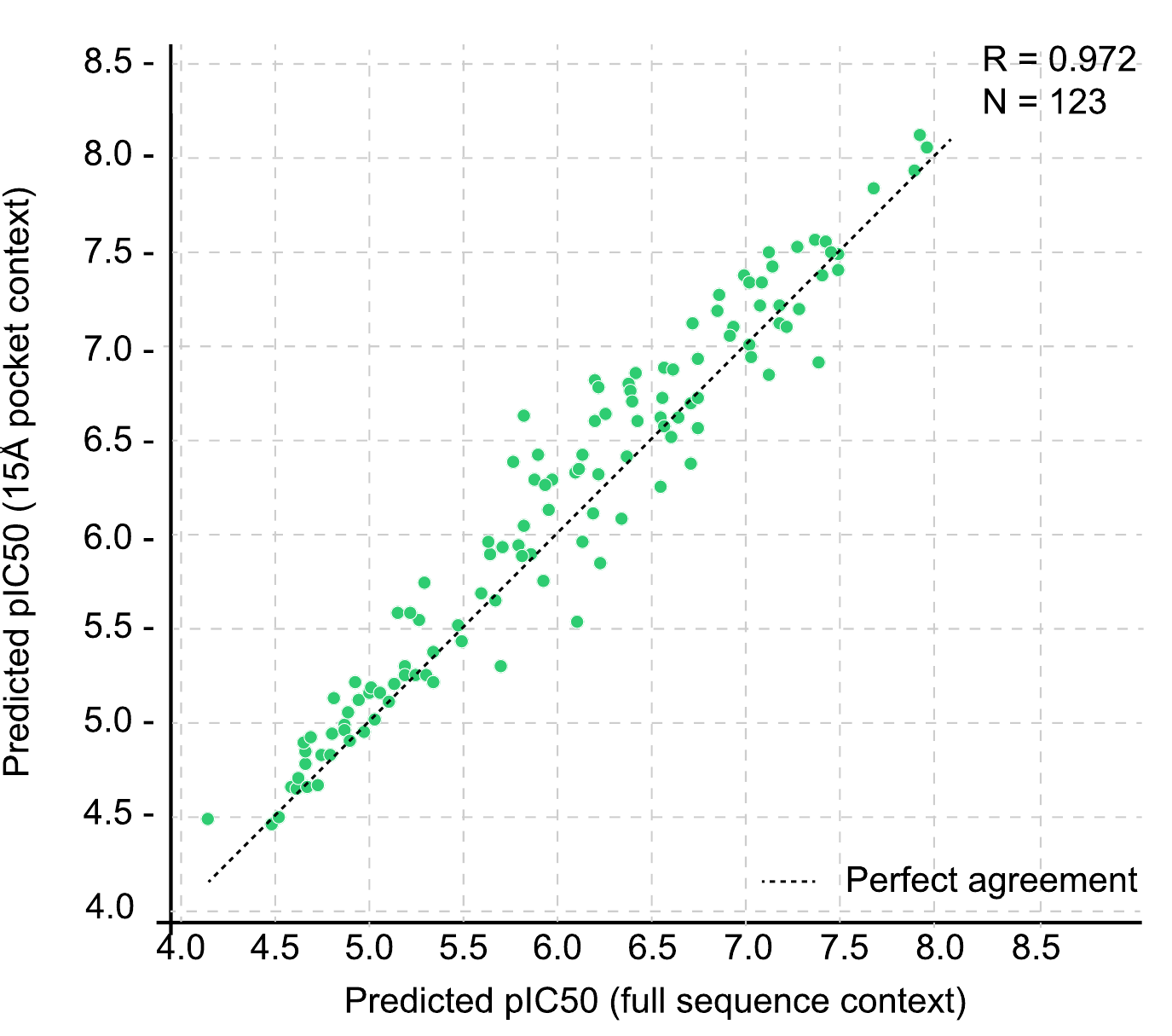}
    \caption{\textbf{Efficiency of limited binding site context on CASP16.} Performance comparison on the full L3000 dataset. The binding site context was defined via the predicted structure of the largest ligand, resulting in a 128-token pocket subset applied to all complexes. This minimal context achieves a $0.95$ correlation with full-sequence results and delivers a $17.16\times$ to $20.3\times$ speedup (single A6000 GPU) relative to the full 846-residue baseline.}
    \label{fig:pocket_vs_nonpocket_affinity}
\end{figure}

In Figure~\ref{fig:pocket_vs_nonpocket_affinity}, we evaluate affinity predictions for the full set of CASP16 L3000 complexes. We adopt a ``minimal token budget'' strategy by defining a single binding site context derived from the largest ligand in the set. Specifically, we perform one full-sequence structural prediction for this ligand and identify protein residues with expected distances less than $15\text{\AA}$ from predicted ligand atoms. This process yielded a 128-residue context, which was subsequently used as input for structure and affinity inference across all remaining molecules. This approach relies entirely on predicted structures. Despite reducing the input from the full 846-residue sequence to just 128 tokens, we observe a $0.95$ correlation with full-context predictions. Benchmarking on a single NVIDIA A6000 GPU demonstrates a $20.3\times$ speedup for affinity prediction ($17.16\times$ with pose generation) while maintaining equivalent accuracy.


\subsection{Sequence-based model baseline}

\begin{figure}[H]
    \centering
    \includegraphics[width=0.75\linewidth]{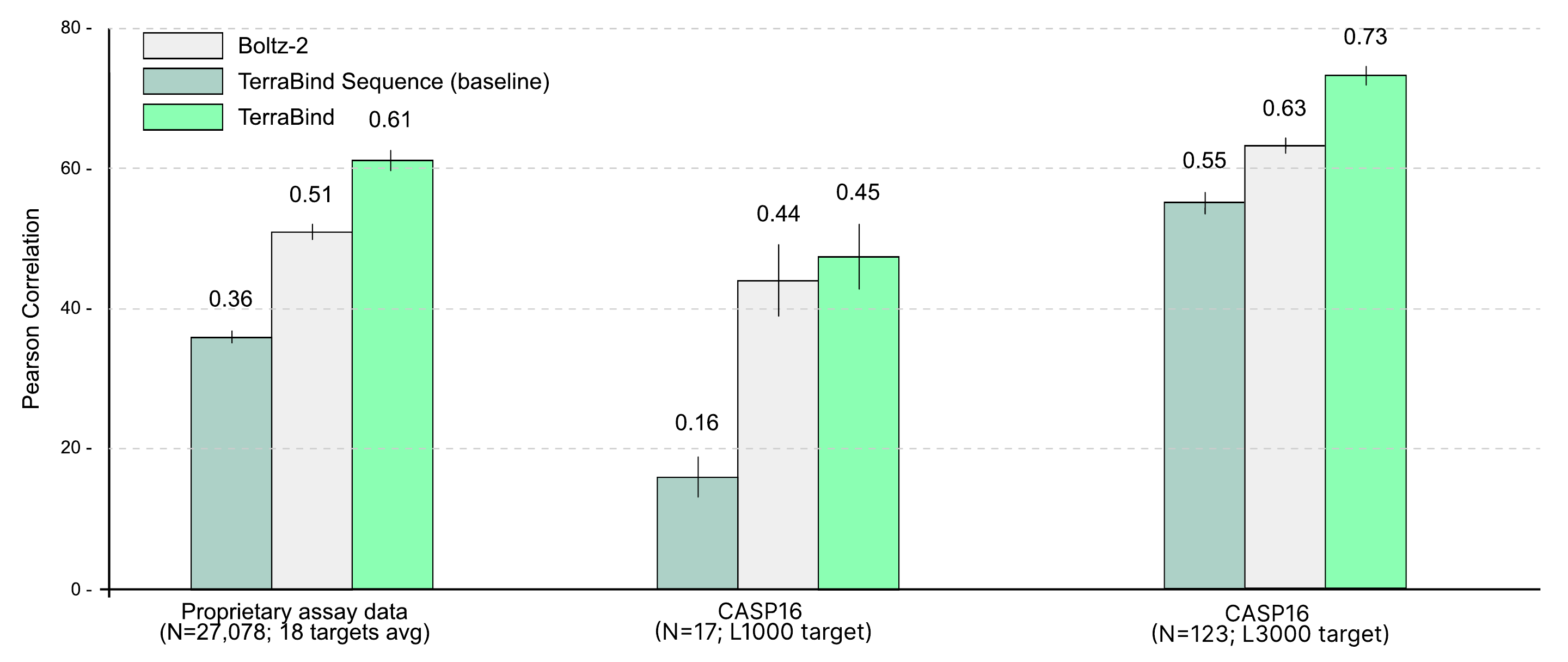}
    \caption{\textbf{Sequence-based model baseline.} A 5M parameter MLP architecture using ESM-2 and the same aggregate SMILES ligand embedding as TerraBind without structural information. TerraBind's structure-aware approach substantially outperforms this baseline.}
    \label{fig:affinity_w_sequence_baseline}
\end{figure}

In Fig.~\ref{fig:affinity_w_sequence_baseline}, we plot the performance of a straightforward \textit{sequence-based} model architecture to serve as a model baseline. This 5M parameter model is constructed by a series of MLPs. Two input MLPs are used to independently process the end-of-sequence ESM-2 residue embedding for the sequence and the COATI-3 ligand embedding. The output embeddings are concatenated and provided to another MLP before diverging at the quantitative and binary output prediction heads. All other aspects of the training data and training process are identical to the process described in Section~\ref{sec:affinity_module_training_procedure}.

\subsection{Binding Pocket Context Size Distribution}

\begin{figure}[H]
    \centering
    \includegraphics[width=0.37\linewidth]{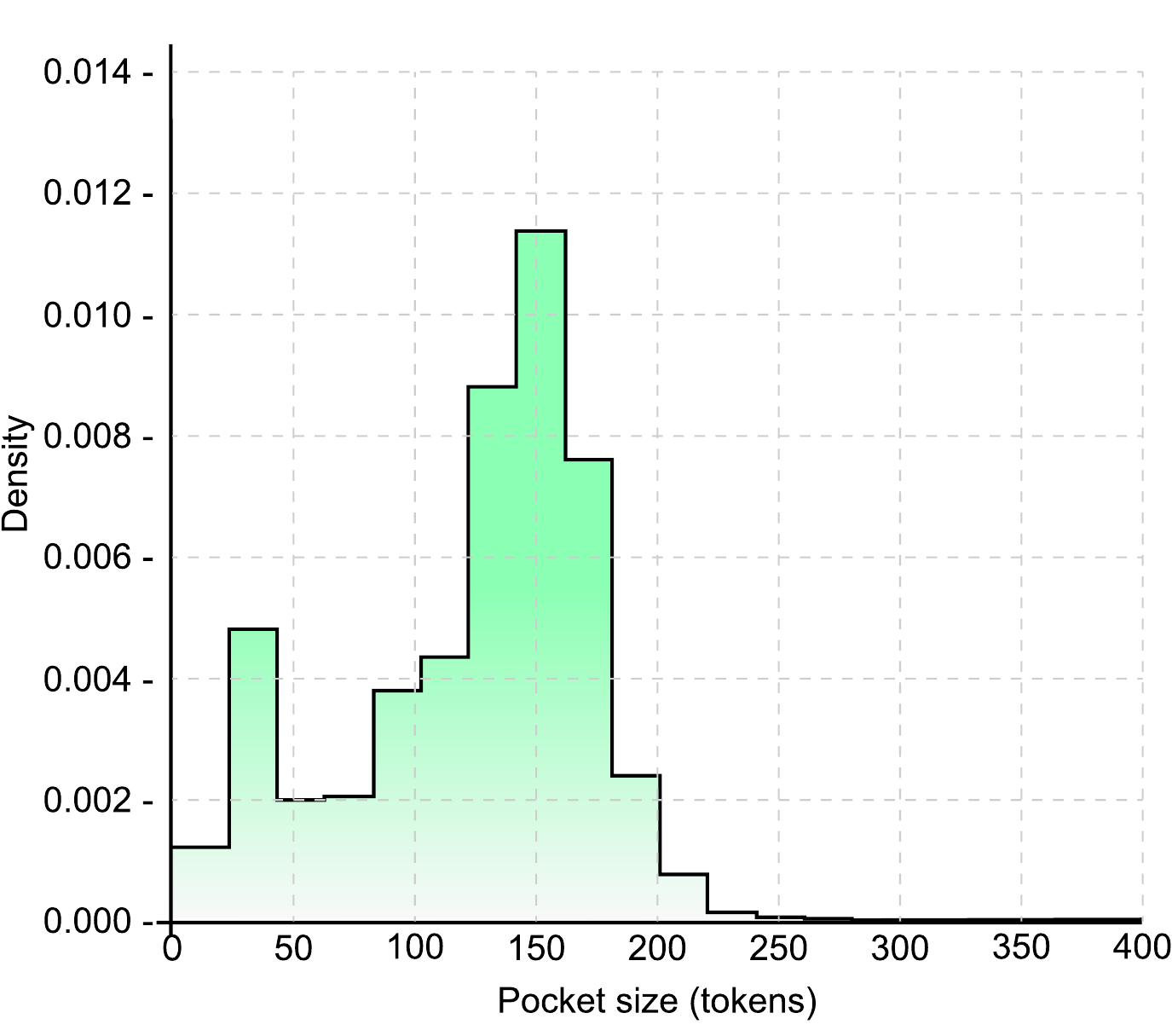}
    \caption{\textbf{Binding pocket context size distribution.} Histogram of token count ($N_{\text{token}}$ = ligand heavy atoms + pocket residues within $15\text{\AA}$) aggregated over the binding affinity dataset. The distribution peaks around 100–150 tokens, validating our 256-token crop size for training and 196-token inference context for TerraBind Pocket.}
    \label{fig:affinity_15A_context}
\end{figure}

Figure~\ref{fig:affinity_15A_context} shows the distribution of binding pocket context sizes across the affinity training dataset. For each protein-ligand complex, the pocket context is defined as the number of ligand heavy atoms plus protein residues predicted to be within $15\text{\AA}$ of any ligand atom, based on expected distances from the pairformer's distogram (Eq.~\ref{eq:expected_pair_distance}). The distribution peaks around 150 tokens, with the vast majority of small-molecule binding contexts requiring fewer than 200 tokens. This empirical observation validates our architectural choices: the 256-token crop size used in Stages 2 and 3 of training, and the 196-token inference context for TerraBind Pocket, are sufficient to capture the full binding site for nearly all drug-like protein-ligand complexes.

\clearpage
\bibliographystyle{plain}
\bibliography{references}

\end{document}